\documentclass[journal]{IEEEtran}
\usepackage{forest}
\usepackage{textgreek}
\usepackage{booktabs}
\usepackage{siunitx}
\usepackage{everyshi}
\usepackage{colortbl}
\usepackage{multirow}
\usepackage{threeparttable}
\usepackage{rotating} 
\usepackage{tabularx}
\usepackage{caption}
\usepackage{hyperref}
\ifCLASSINFOpdf
\else
\fi
\begin{document}
%
\title{Data Augmentation for Time-Series Classification: a Comprehensive Survey}
%
%
%

\title{Data Augmentation for Time-Series Classification: An Extensive Empirical Study \\and Comprehensive Survey}

\author{Zijun~Gao, Haibao Liu and Lingbo~Li
\thanks{Zijun Gao is with the School of Engineering, University of Warwick, CV4
7AL Coventry, U.K. (e-mail: zijun.gao.1@warwick.ac.uk).}

\thanks{Haibao Liu is with the School of Engineering and Materials Science,
       Queen Mary University of London, London E1 4NS, U.K. (e-mail: haibao.liu@qmul.ac.uk).}
       
\thanks{Lingbo Li is with the School of Engineering, University of Warwick, CV4
7AL Coventry, U.K., State Key Laboratory for Novel Software Technology, Nanjing University, China, and also with Turing Intelligence Technology Limited, London EC2Y 9AW, U.K. (e-mail: Lingbo.Li.1@warwick.ac.uk and lingbo@turintech.ai).}
}

\maketitle

\begin{abstract}
Data Augmentation (DA) has become a critical approach in Time Series Classification (TSC), primarily for its capacity to expand training datasets, enhance model robustness, introduce diversity, and reduce overfitting. 
However, the current landscape of DA in TSC is plagued with fragmented literature reviews, nebulous methodological taxonomies, inadequate evaluative measures, and a dearth of accessible and user-oriented tools. 
This study addresses these challenges through a comprehensive examination of DA methodologies within the TSC domain.
Our research began with an extensive literature review spanning a decade, revealing significant gaps in existing surveys and necessitating a detailed analysis of over 100 scholarly articles to identify more than 60 distinct DA techniques. 
This rigorous review led to the development of a novel taxonomy tailored to the specific needs of DA in TSC, categorizing techniques into five primary categories: Transformation-Based, Pattern-Based, Generative, Decomposition-Based, and Automated Data Augmentation.
This taxonomy is intended to guide researchers in selecting appropriate methods with greater clarity.
In response to the lack of comprehensive evaluations of foundational DA techniques, we conducted a thorough empirical study, testing nearly 20 DA strategies across 15 diverse datasets representing all types within the UCR time-series repository. 
Using ResNet and LSTM architectures, we employed a multifaceted evaluation approach, including metrics such as Accuracy, Method Ranking, and Residual Analysis, resulting in a benchmark accuracy of 84.98 ± 16.41\% in ResNet and 82.41 ± 18.71\% in LSTM.
Our investigation underscored the inconsistent efficacies of DA techniques, for instance, methods like RGWs and Random Permutation significantly improved model performance, whereas others, like EMD, were less effective.
Furthermore, we found that the intrinsic characteristics of datasets significantly influence the success of DA methods, leading to targeted recommendations based on empirical evidence to help practitioners select the most suitable DA techniques for specific datasets.
To improve practical use, we have consolidated most of these methods into a unified Python Library, whose user-friendly interface facilitates experimenting with various augmentation techniques, offering practitioners and researchers a more convenient tool for innovation than currently available options.
\end{abstract}

\begin{IEEEkeywords}
Data Augmentation, Time-Series Classification, Machine Learning, Deep Learning.
\end{IEEEkeywords}

%
\IEEEpeerreviewmaketitle

\section{Introduction}
%
%
%
%
\IEEEPARstart{I}{n} machine learning, the pursuit of robust model performance often depends on the availability of extensive datasets, a challenge exacerbated by the frequent scarcity of data \cite{shorten2019survey}.
Data augmentation addresses this issue by generating additional, analogous data to expand limited datasets, thereby mitigating data insufficiency constraints. 
This technique, renowned for its cost-effectiveness and efficacy, has gained immense traction in various machine learning disciplines, significantly improving model accuracy in environments characterized by limited data.
For instance, in computer vision, image augmentation is relatively straightforward, with a variety of simple yet effective techniques capable of producing discernible images even when subjected to noise or partial cropping \cite{jung2017imgaug}.

In contrast, time-series data augmentation has received less attention. 
Existing strategies often overlook the intrinsic properties defining time-series data, such as temporal correlation, a critical aspect that distinguishes it from other data types.
The potential for transforming time-series data into frequency or time-frequency domains further complicates the application of conventional augmentation techniques, particularly those borrowed from image or speech processing, leading to potential inaccuracies in synthetic data generation.
Additionally, the effectiveness of data augmentation methods is highly task-dependent, requiring a nuanced approach that addresses the unique demands of tasks like forecasting and classification within the time-series context.

This paper offers an extensive examination of the current state of Data Augmentation (DA) for Time Series Classification (TSC), exploring a wide range of techniques and methodologies that have shaped the field over the past decade, providing readers with a comprehensive understanding of its evolution.

\textbf{- Comprehensive Survey and Novel Taxonomy:} To bring clarity and structure to the diverse array of techniques in this field, we introduce a novel taxonomy specifically designed for DA in TSC.
This classification system organizes methods into five fundamental categories: Transformation-Based, Pattern-Based, Generative, Decomposition-Based, and Automated Data Augmentation, enabling a systematic and streamlined approach to selecting and analyzing DA methods for TSC.

\textbf{- Detailed Evaluation:} Beyond a traditional survey and taxonomy, this study makes a significant contribution by rigorously evaluating nearly twenty DA techniques for TSC. 
These methods were tested using two established neural network models, providing a robust performance benchmark. 
Our empirical analysis offers a comprehensive perspective on the relative strengths and weaknesses of these techniques, delivering valuable insights for both academia and industry.

\textbf{- Unified Library:} To complement our efforts, we have developed a Python Library that encapsulates most of the aforementioned time-series data augmentation algorithms, designed for ease of implementation\footnote{https://github.com/RangerG/Dataugmenter}. 
This initiative addresses the shortcomings of existing tools, offering a more user-friendly alternative. 

The insights gained from our research provide a broad view of current techniques while highlighting potential areas for improvement and future research.
We aim to foster further innovation in DA for TSC, paving the way for the development of more sophisticated and effective methods tailored to the complexities of time-series data.
The main contributions of this manuscript are as follows:\\

\begin{itemize}
    \item We conduct a thorough review of advancements in Data Augmentation for Time Series Classification (DA for TSC) over the past decade;

    \item We advocate a novel, purpose-built taxonomy for DA for TSC, segmenting techniques into quintessential categories: Transformation-Based, Pattern-Based, Generative, Decomposition-Based, and Automated Data Augmentation;

    \item We perform an in-depth evaluation of nearly twenty DA for TSC techniques, utilizing two renowned neural network models, and offer an exhaustive discourse on the findings;

    \item We introduce a Python Library of preeminent time series data augmentation algorithms, offering a unified, user-friendly platform for the application of diverse data augmentation strategies to time series datasets.\\
\end{itemize}

\begin{figure*}[ht]
    \centering
    \includegraphics[scale=0.2]{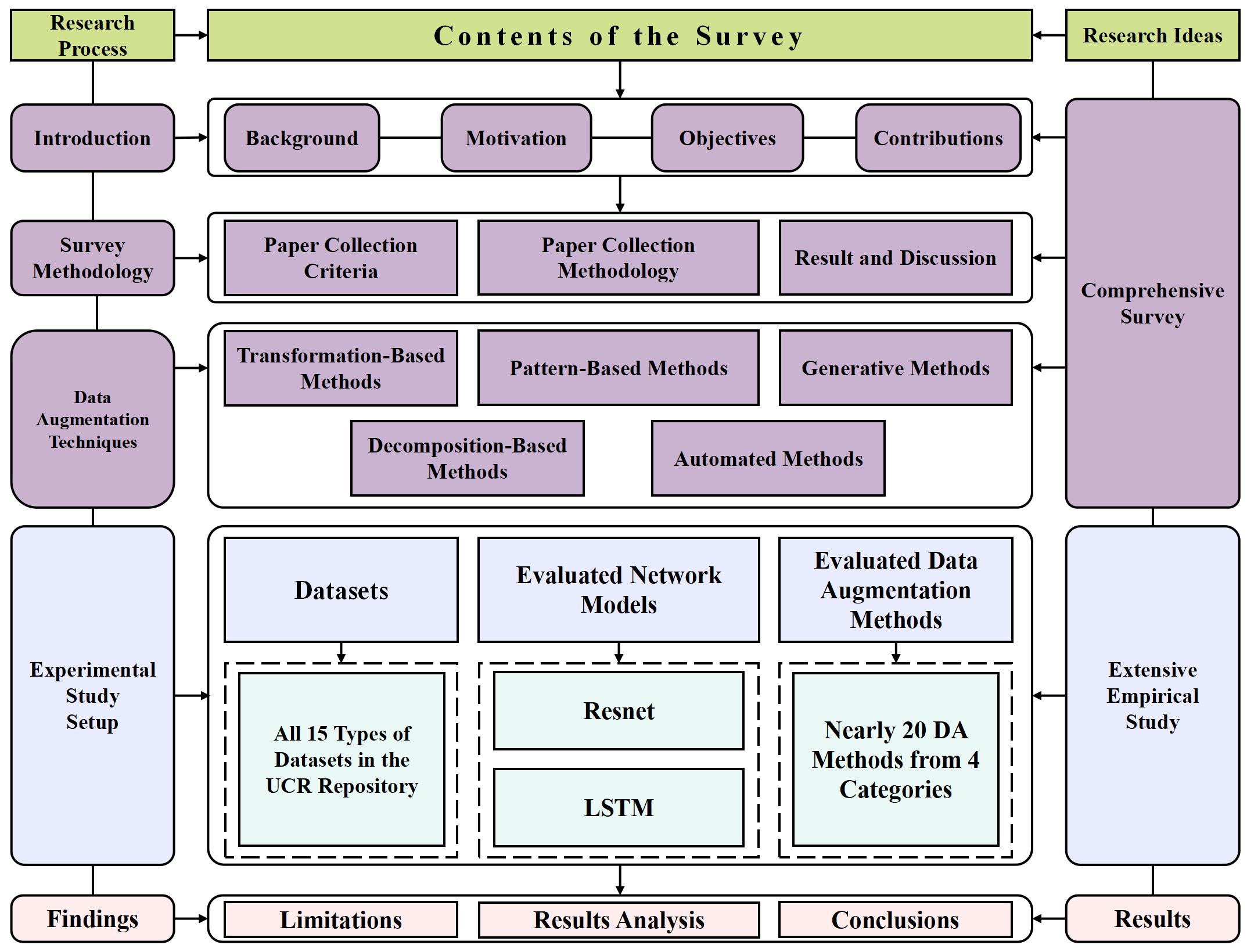}
    \caption{The Overall Organization of this Survey Paper}
    \label{fig:Organization_of_paper}
\end{figure*}

The structure of this article is meticulously organized to facilitate a comprehensive understanding of our investigation, as depicted in Figure \ref{fig:Organization_of_paper}. Section \ref{Literature Review} embarks on a thorough literature review, delineating the state of the art in Data Augmentation (DA) for Time Series Classification (TSC), establishing the foundation upon which our study is predicated. 
In Section \ref{Survey Methodology}, we elaborate on the methodological framework adopted for our survey, including the criteria for selection and analysis of relevant works, and engage in a detailed discussion of the findings derived therefrom.
Section \ref{Data Augmentation for Time-Series Classification Techniques} introduces a novel taxonomy for DA techniques, methodically categorizing these methodologies into five distinct principal categories, thereby offering a structured lens through which the current landscape can be viewed. 
Section \ref{Experimental Study Setup} is dedicated to the elucidation of our experimental design, meticulously detailing the setup employed in our empirical study and articulating the research questions poised at the outset.
Subsequently, Section \ref{Results Analysis} addresses these research questions, presenting a critical analysis of the experimental outcomes and drawing insights pertinent to the field of DA for TSC. 
Section \ref{Limitations and Threats to Validity} ventures into an exploration of the limitations inherent to our study and the potential threats to the validity of our findings, fostering a transparent discourse on the scope and applicability of our conclusions. 
The article culminates in Section \ref{Conclusions and Prospects for Future Research}, where we synthesize our contributions to the domain and outline directions for future research.

\section{Background and Fundamentals}
\label{Literature Review}

\textbf{Time Series Classification:} 
The domain of Time Series Classification (TSC) represents a pivotal frontier in data mining, dedicated to the categorization of time series into distinct classes predicated upon historical pattern analysis \cite{forestier2019deep}.
Inherent in a time series is a sequence of data points, chronologically ordered and indexed over regular intervals, underscoring the temporal dynamics integral to various practical realms, including health diagnostics, financial market analytics, among others.
The intricate nature of time series data, marked by heterogeneous lengths, noise interferences, and nonlinear trajectories, poses substantial analytical challenges.

\textbf{Data Augmentation:} 
Within the spheres of machine learning and deep learning, data augmentation stands out as a strategy for amplifying the training dataset by generating modified iterations of the original data instances \cite{iwana2021empirical}.
This process, applicable across diverse tasks such as image recognition, involves transformations such as rotation, scaling, or translation.
In the context of TSC, augmentation encompasses techniques like noise injection, slicing, jittering, and scaling, serving dual core objectives: enhancing model generalization to diminish overfitting and addressing class imbalances within datasets.

Despite its established efficacy in various machine learning applications, data augmentation's integration into TSC remains a vibrant research avenue, necessitating deeper exploration.
Historically, data augmentation has proven indispensable in scenarios plagued by data scarcity or imbalance, as evidenced in applications like AlexNet for ImageNet classification \cite{krizhevsky2012imagenet}.
Recent scholarly endeavors have extended these techniques to time series data, albeit with modifications to accommodate the data's unique characteristics \cite{um2017data,rashid2019window,bishop1995training,an1996effects}.

The singular nature of time series data often precludes the direct application of standard augmentation methods, necessitating adaptations or entirely novel methodologies, particularly for tasks like classification, forecasting, and anomaly detection within time series \cite{talavera2022data}.
The past decade has witnessed a proliferation of TSC algorithms, spurred by the increased availability of time-series data and its ubiquity across tasks necessitating human cognitive processes \cite{bagnall2017great,gamboa2017deep}.
The diverse dataset types within the UCR/UEA archive further testify to the expansive applicability of TSC \cite{dau2019ucr}.

Addressing data imbalances, a prevalent issue in machine learning tasks, requires strategic data augmentation interventions, such as oversampling or undersampling, to rectify biases towards majority classes and enhance minority class performance.
Techniques like SMOTE (Synthetic Minority Over-sampling Technique) generate synthetic examples to balance class distributions, while their counterparts reduce majority class instances \cite{li2021shapelets}.
However, these approaches necessitate careful calibration to avoid overfitting or information loss.

Innovative contributions to the field include Guennec et al.'s two-layer convolutional network for TSC \cite{le2016data}, Germain et al.'s time series-specific augmentation methods \cite{forestier2017generating}, and Pfister et al.'s exploration of wearable sensor data augmentation \cite{um2017data}.
Haradal et al. ventured into biosignal classification, employing generative adversarial networks (GAN) for synthetic time series data generation \cite{haradal2018biosignal}.
Fons et al. examined financial time series classification, evaluating various data augmentation strategies with deep learning models \cite{fons2020evaluating}.
Nonaka et al. proposed ECG-specific data augmentation, enhancing atrial fibrillation classification using single-lead ECG data without modifying the DNN architecture \cite{nonaka2020data}.
Goubeaud et al. introduced White Noise Windows, a novel time series data augmentation technique, in 2021 \cite{goubeaud2021white}.

Despite its advantages, data augmentation is not without its challenges, including the risk of overfitting and the potential ineffectiveness with high-dimensional or complex data structures \cite{iwana2021empirical,wen2020time}.

\subsection{Relevant Survey \& Study}
The scholarly landscape is replete with surveys on data augmentation, spanning domains from image processing \cite{naveed2021survey} to natural language processing \cite{feng2021survey}, though comprehensive studies in the time series sector remain scarce.
Iwana et al. conducted an empirical assessment of time series data augmentation methods using neural networks, albeit with a focus limited to classification and excluding frequency domain or externally trained methods \cite{iwana2021empirical}.
Wen et al. expanded upon this work, incorporating time series forecasting and anomaly detection, though their study utilized a relatively constrained dataset from the Alibaba Cloud monitoring system \cite{wen2020time}.

Although these studies offer valuable insights into data augmentation techniques, there is a conspicuous absence of a holistic view that encompasses recent advances and diverse application fields.
Several data augmentation techniques referenced in the literature are either antiquated or no longer align with contemporary research and practice standards, thereby failing to offer a user-friendly and accessible toolset for researchers and practitioners. 
This article endeavors to address this gap in the scholarly discourse by conducting a comprehensive analysis of the current landscape of data augmentation within the domain of time-series classification. 
It aims to serve as a pivotal reference point, guiding future research directions and fostering innovation in this rapidly evolving field.

\section{Survey Methodology}
\label{Survey Methodology}

This paper adopts a tripartite research methodology encompassing Survey, Evaluation, and Experimental Study Setup. 
The Survey delineates our systematic approach to literature accumulation and analysis. 
The Evaluation introduces a novel taxonomy of time series data augmentation, predicated on distinctive characteristics. 
The Experimental Study Setup discusses the authoritative datasets employed, evaluation metrics, and potential variables influencing outcomes, thereby ensuring a rigorous analysis of time series data augmentation methodologies.

\subsection{Paper Collection Criteria of the Survey}

The literature aggregation for this study adheres to stringent inclusion criteria, encompassing:

\begin{itemize}
  \item Papers elucidating the fundamental concepts of data augmentation for time-series classification or its associated facets.
  \item Studies proposing enhancements to existing data augmentation methodologies, tools, or frameworks within the realm of time-series classification.
  \item Contributions introducing benchmarks or datasets tailored for data augmentation in time-series classification.
  \item Literature offering comprehensive reviews or surveys on the contemporary landscape of data augmentation for time-series classification.
\end{itemize}

Exclusions pertain to studies applying relevant techniques outside the time-series context and those focusing on time-series classification, forecasting, or anomaly detection without integrating data augmentation methodologies.

\subsection{Paper Collection Methodology of the Survey}

Our literature retrieval process commenced with exhaustive searches across renowned scientific repositories like Google Scholar and arXiv, utilizing precise keyword combinations. 
Recognizing the interdisciplinary application of time series data augmentation, we adopted a snowballing strategy for comprehensive coverage \cite{wohlin2014guidelines}. 
The accumulated literature was subsequently categorized as Survey/Review papers, Research papers, or Methods papers, following established academic conventions \cite{booth2003craft,creswell2017research,gastel2022write}.

The keywords used for searching and collecting are listed below. [DA for
TSC] means Data Augmentation for Time-Series Classification, which is the focus of
this research project.
The search strategy encompassed several iterations, as detailed below:

\begin{itemize}
  \item DA for TSC + Survey / Review
  \item DA for TSC + Techniques / Methods
  \item DA for TSC + Evaluations / Comparison
\end{itemize}

\subsection{Result and Discussion of the Survey}

\begin{table}[!t]
\renewcommand{\arraystretch}{1.3}
\caption{Paper Query Results}
\label{tab:Paper_Query_Results}
\centering
\scalebox{0.95}{
\begin{tabular}{|l|c|c|c|}
\hline
\textbf{Paper Query Results} & \textbf{Key Words} & \textbf{Hits} & \textbf{Title} \\ \hline
[DA for TSC] + Survey / Review & 25 & 20 & 8 \\ \hline
[DA for TSC] + Techniques / Methods & 273 & 102 & 72 \\ \hline
[DA for TSC] + Evaluations / Comparison & 90 & 55 & 26 \\ \hline
Query & - & - & 106 \\ \hline
Snowball & - & - & 41 \\ \hline
Overall & - & - & 147 \\ \hline
\end{tabular}
}
\end{table}

\begin{figure}[!t]
    \centering
    \includegraphics[scale=0.2]{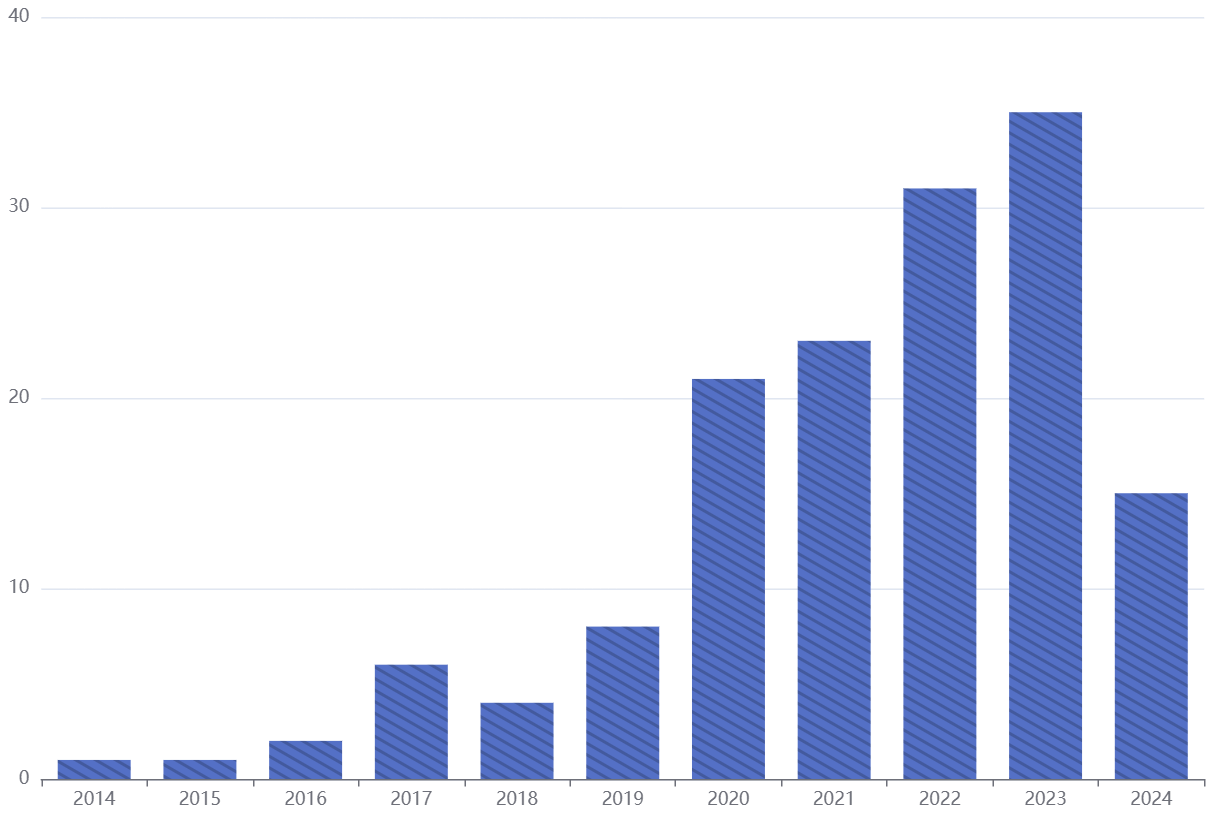}
    \caption{Annual Publication Trends in Data Augmentation for Time-Series Classification}
    \label{fig:number_of_paper}
\end{figure}

The terms ``Key Words", ``Hits” and ``Title” refer to the number of papers returned by the search, the number of papers that remain after title framing and the number of papers that remain after body framing, respectively.
The search strategy yielded 147 relevant publications until 10th June, 2024, with a significant proportion originating from keyword searches and the remainder from snowballing. 
Trends in annual publication indicate a growing interest in this research area, particularly after 2019, as shown in Figure \ref{fig:number_of_paper}.

The distribution of publications across various publishers underscores the dominance of IEEE and arXiv, with significant contributions also emanating from Elsevier, MDPI, ACM, and Springer.

\begin{figure}[!t]
    \centering
    \includegraphics[scale=0.2]{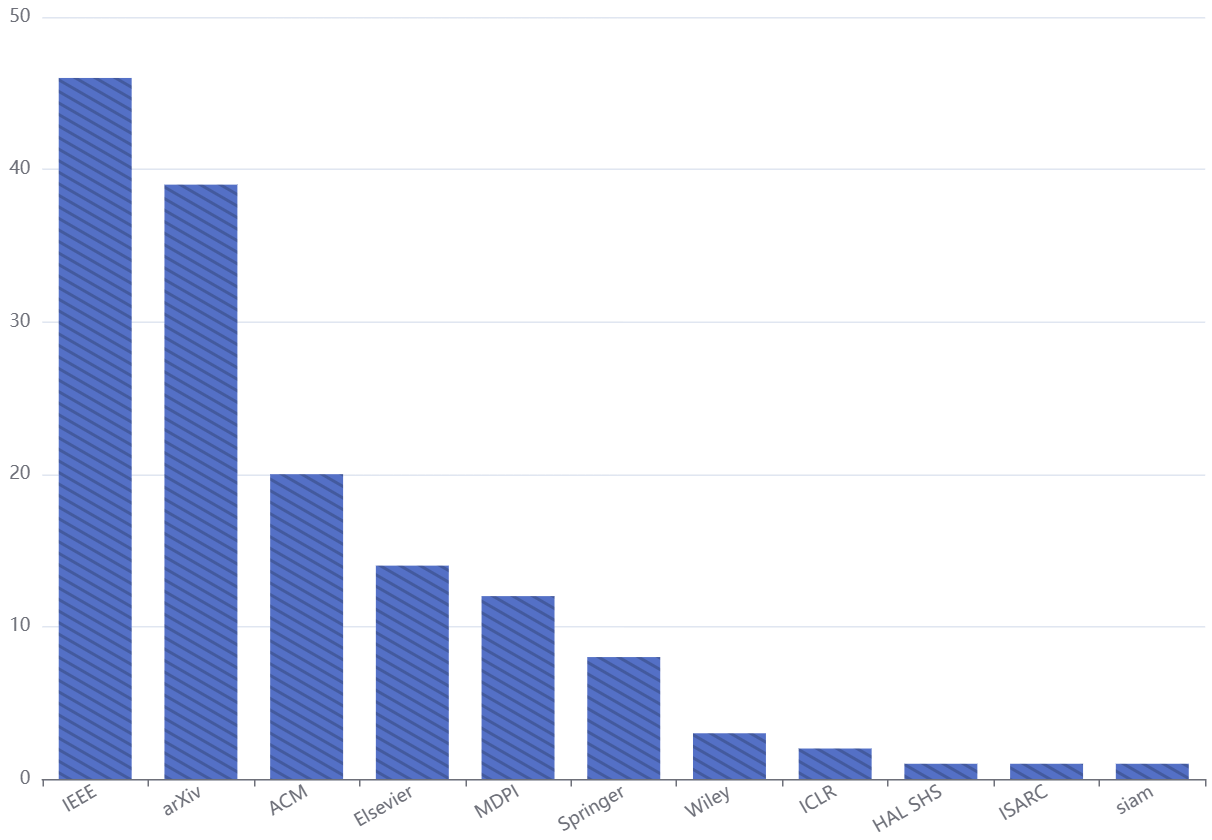}
    \caption{Publisher Distribution of Publications in Data Augmentation for Time-Series Classification}
    \label{fig:by_publisher}
\end{figure}

A geographical analysis of the publications reveals a preponderance of contributions from the United States, followed by China, Japan, Germany, and the United Kingdom, with disparate representation from other global regions.

\begin{figure}[!t]
    \centering
    \includegraphics[scale=0.18]{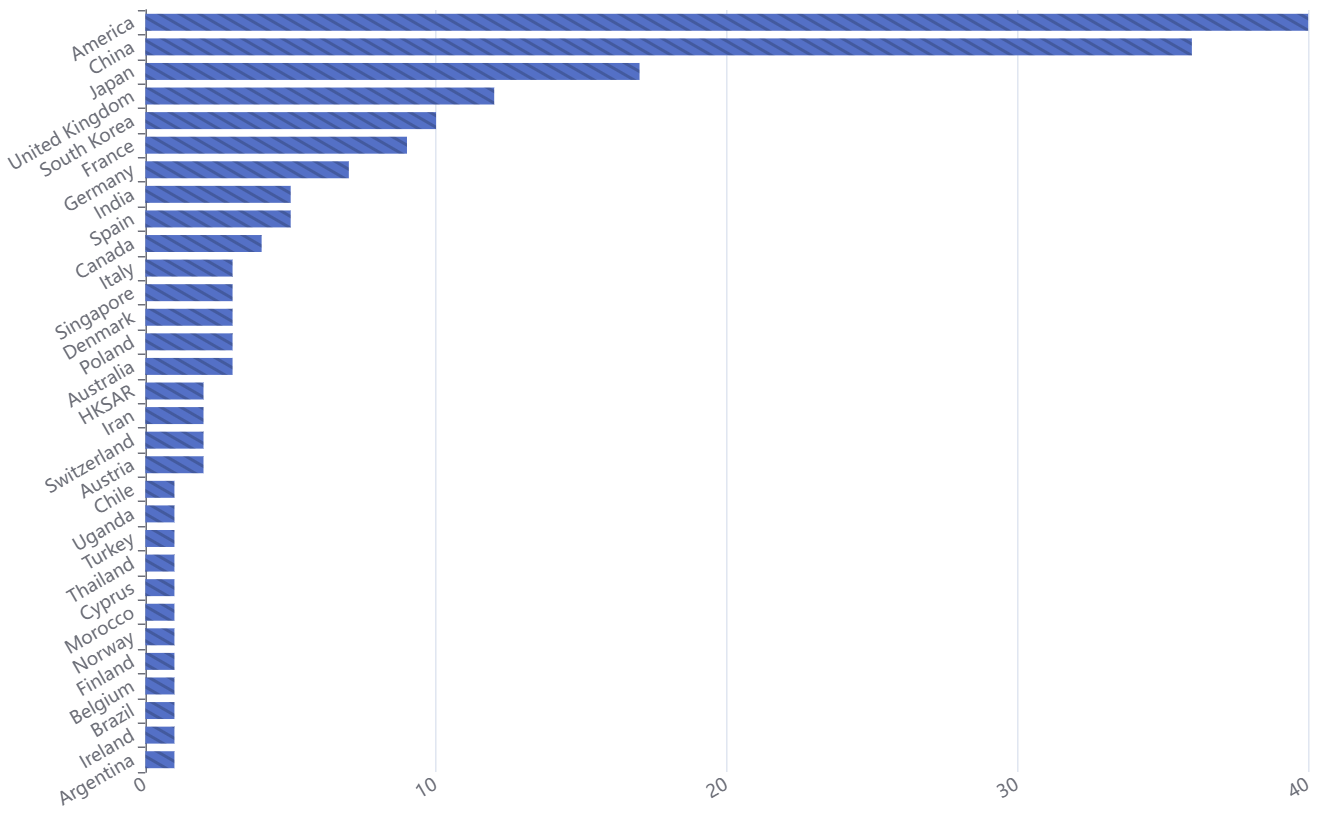}
    \caption{Publication Count by Country in Data Augmentation for Time-Series Classification}
    \label{fig:by_country}
\end{figure}

An industry-centric examination, based on the North American Industry Classification System (NAICS), highlights a predominant application within the Information Technology sector, followed by Healthcare, Agriculture, Astronomy, Finance, and Construction, indicating the methodologies' broad applicability.

\begin{figure}[!t]
    \centering
    \includegraphics[scale=0.2]{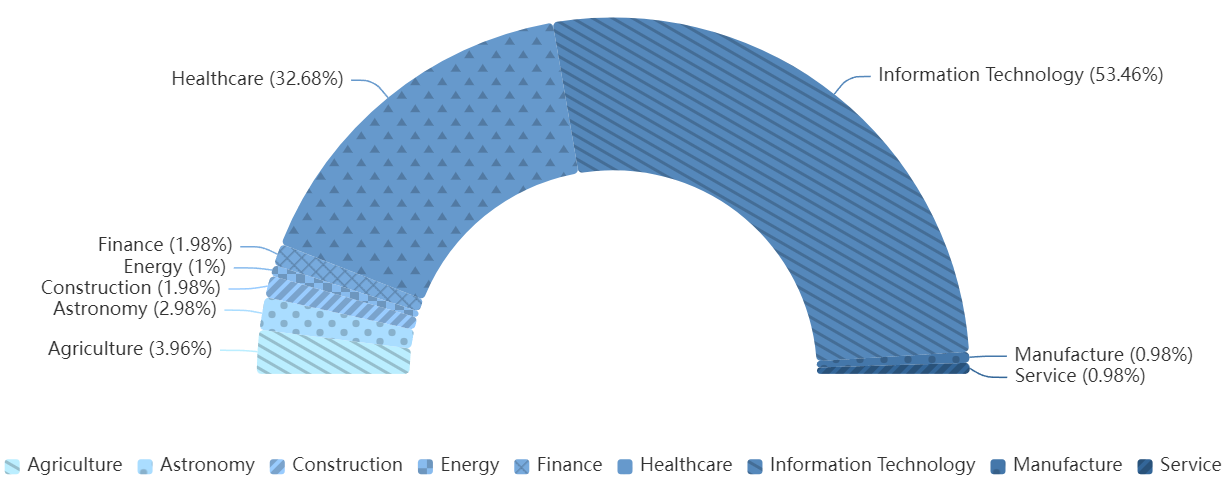}
    \caption{Publication Count by Industry in Data Augmentation for Time-Series Classification}
    \label{fig:by_field}
\end{figure}

\section{Data Augmentation for Time-Series Classification Techniques}
\label{Data Augmentation for Time-Series Classification Techniques}

The taxonomy presented in this study for data augmentation techniques applicable to time-series classification (TSC) emerges from an exhaustive literature survey and rigorous research.
It compartmentalizes the techniques into five principal categories: Transformation-Based Methods, Pattern-Based Methods, Generative Methods, Decomposition-Based Methods, and Automated Data Augmentation Methods.
This categorization, meticulously refined and expanded upon from preceding studies \cite{talavera2022data,iwana2021empirical,wen2020time}, lays the groundwork for our in-depth exploration and experimentation.

Several factors underpin the formulation of this taxonomy.
Primarily, it encompasses a broad spectrum of contemporary methods, thereby paving the way for domain-centric experimental undertakings in specialized segments like the time and frequency domains.
Moreover, it offers an organized synopsis of the myriad techniques used for augmenting time-series data, thereby simplifying the comparative analysis and comprehension of diverse methods.
Such a structured approach is indispensable for researchers and practitioners alike, guiding them in the judicious selection of data augmentation techniques tailored to their specific requirements.
Furthermore, by highlighting the existing lacunae and prospects in the current scenario, it spurs the innovation of novel techniques.

The distinct categories within this taxonomy epitomize different methodologies in data augmentation.
Transformation-Based Methods involve the alteration of original data via various transformations.
Pattern-Based Methods, on the other hand, generate new instances through the extraction and subsequent recombination of patterns inherent in the original data.
Generative Methods utilize generative models to synthesize new data instances.
Decomposition-Based Methods, in contrast, entail the disintegration of the original time-series into multiple components, which are then individually augmented and reassembled.
Automated Data Augmentation Methods streamline the augmentation process, often leveraging data-driven insights to ascertain the most efficacious augmentation strategy.

The ensuing subsections provide an exhaustive discourse on each category, elucidating their respective methodologies, merits, and potential use-cases.
This detailed analysis serves dual purposes: guiding users in the selection of appropriate data augmentation strategies and fostering the conception of avant-garde methods in forthcoming research endeavors.

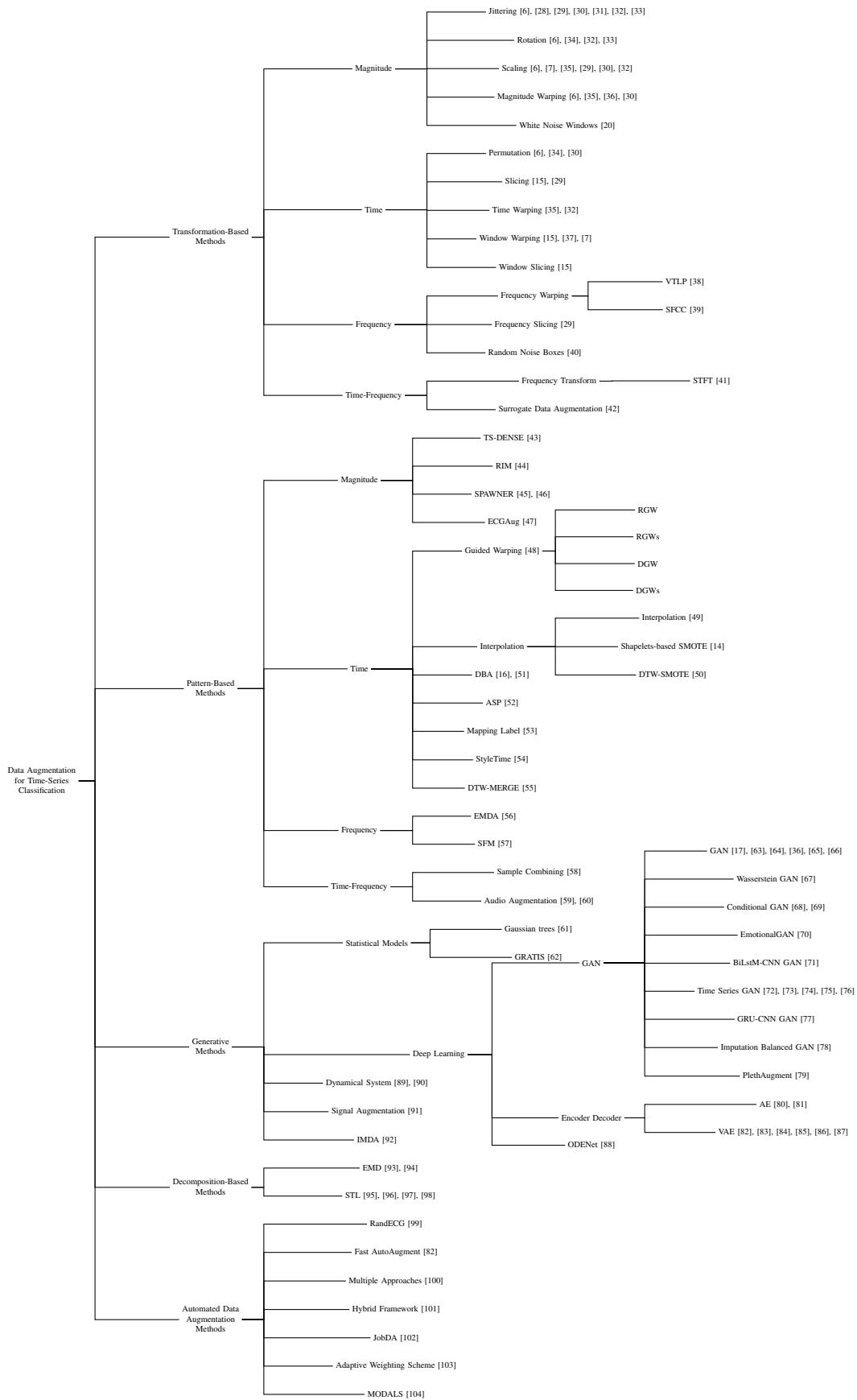
\begin{figure*}[!t]
\hypersetup{hidelinks}  
\centering
\captionsetup{justification=centering}  
\begin{forest}
for tree={
    grow'=east,
    s sep=7pt, 
    l sep=1.5cm, 
    anchor=base,
    edge path={
        \noexpand\path[\forestoption{edge}]
        (!u.parent anchor) -- +(25pt,0) |- (.child anchor)\forestoption{edge label};
    },
    scale=0.4, 
    transform shape, 
    font=\normalsize, 
}
[Data Augmentation\\for Time-Series\\Classification, align=center, before computing xy={l=2cm}
    [Transformation-Based\\Methods, align=center
        [Magnitude
            [Jittering \cite{um2017data,zhang2017perspective,cai2022performance,liu2020efficient,kim2020deep,rashid2019times,huang2019data}]
            [Rotation \cite{um2017data,li2021data,rashid2019times,huang2019data}]
            [Scaling \cite{um2017data,rashid2019window,pi2022improving,cai2022performance,liu2020efficient,rashid2019times}]
            [Magnitude Warping \cite{um2017data,pi2022improving,do2022data,liu2020efficient}]
            [White Noise Windows \cite{goubeaud2021white}]
        ]
        [Time
            [Permutation \cite{um2017data,li2021data,liu2020efficient}]
            [Slicing \cite{le2016data,cai2022performance}]
            [Time Warping \cite{pi2022improving,rashid2019times}]
            [Window Warping \cite{le2016data,warchol2022efficient,rashid2019window}]
            [Window Slicing \cite{le2016data}]
        ]
        [Frequency
            [Frequency Warping
                [VTLP \cite{jaitly2013vocal}]
                [SFCC \cite{yang2023sfcc}]
            ]
            [Frequency Slicing \cite{cai2022performance}]
            [Random Noise Boxes \cite{goubeaud2021random}]
        ]
        [Time-Frequency
            [Frequency Transform
                [STFT \cite{steven2018feature}]
            ]
            [Surrogate Data Augmentation \cite{lee2019surrogate}]
        ]
    ]
    [Pattern-Based\\Methods, align=center
            [Magnitude
            [TS-DENSE \cite{zanella2022ts}]
            [RIM \cite{aboussalah2022recursive}]
            [SPAWNER \cite{kamycki2019data,warchol2022augmentation}]
            [ECGAug \cite{stabenau2021ecgaug}]
        ]
        [Time
            [Guided Warping \cite{iwana2021time}
                [RGW]
                [RGWs]
                [DGW]
                [DGWs]
                ]
            [Interpolation
            [Interpolation \cite{oh2020time}]
            [Shapelets-based SMOTE \cite{li2021shapelets}]
            [DTW-SMOTE \cite{yang2021time}]
            ]
            [DBA \cite{forestier2017generating,li2022novel}]
            [ASP \cite{liu2022adaptive}]
            [Mapping Label \cite{khalili2021automatic}]
            [StyleTime \cite{el2022styletime}]
            [DTW-MERGE \cite{akyash2021dtwmerge}]
        ]
        [Frequency
            [EMDA \cite{loris2020paci}]
            [SFM \cite{cui2015data}]
        ]
        [Time-Frequency
            [Sample Combining \cite{zhao2020classification}]
            [Audio Augmentation \cite{nanni2020data,park2019specaugment}]
        ]
    ]
    [Generative\\Methods, align=center
        [Statistical Models
            [Gaussian trees \cite{cao2014parsimonious}]
            [GRATIS \cite{kang2020gratis}]
        ]
        [Deep Learning, l=2cm
            [GAN
                [GAN \cite{haradal2018biosignal,hatamian2020effect,garcia2022improving,do2022data,albert2021data,morizet2022pilot}]
                [Wasserstein GAN \cite{lou2018one}]
                [Conditional GAN \cite{ramponi2018t,ehrhart2022conditional}]
                [EmotionalGAN \cite{chen2019emotionalgan}]
                [BiLstM-CNN GAN \cite{gregor2015draw}]
                [Time Series GAN \cite{yang2023ts,li2022tts,smith2021spectral,li2021tsa,smith2020conditional}]
                [GRU-CNN GAN \cite{liu2022time}]
                [Imputation Balanced GAN \cite{deng2022ib}]
                [PlethAugment \cite{kiyasseh2020plethaugment}]
            ]
            [Encoder Decoder
                [AE \cite{zha2022time,devries2017dataset}]
                [VAE \cite{park2022dimensional,alawneh2021enhancing,feng2021intelligent,moreno2020improving,iglesias2023data,hsu2017unsupervised}]
            ]
            [ODENet \cite{sarkar2020neural}]
        ]
        [Dynamical System \cite{minati2023accelerometer,perl2020data}]
        [Signal Augmentation \cite{yeomans2019simulating}]
        [IMDA \cite{wang2019image}]
    ]
    [Decomposition-Based\\Methods, align=center
        [EMD \cite{li2022time,nam2020data}]
        [STL \cite{cleveland1990stl,wen2019robuststl,bergmeir2016bagging,wen2020fast}]
    ]
    [Automated Data\\Augmentation\\Methods, align=center
        [RandECG \cite{nonaka2021randecg}]
        [Fast AutoAugment \cite{park2022dimensional}]
        [Multiple Approaches \cite{li2022integrated}]
        [Hybrid Framework \cite{wu2021non}]
        [JobDA \cite{ma2021joint}]
        [Adaptive Weighting Scheme \cite{fons2021adaptive}]
        [MODALS \cite{cheung2020modals}]
    ]
]
\end{forest}
\caption{Taxonomy of Techniques of Data Augmentation for Time-Series Classification}
\label{fig:mytree}
\hypersetup{hidelinks=false}  
\end{figure*}

\subsection{Transformation-Based Methods}
Transformation-based methods hinge on the principle of applying an array of transformations to the original time-series data. 
The objective is to retain the fundamental structure while introducing nuanced variations, thereby bolstering model resilience. 
These methods, characterized by their simplicity and intuitiveness, typically encompass elementary mathematical operations, making them versatile across diverse data types and domains. 
The scope of transformations extends to various data facets, culminating in four distinct sub-categories.

\begin{figure}[!t]
    \centering
    \includegraphics[width=3.5in]{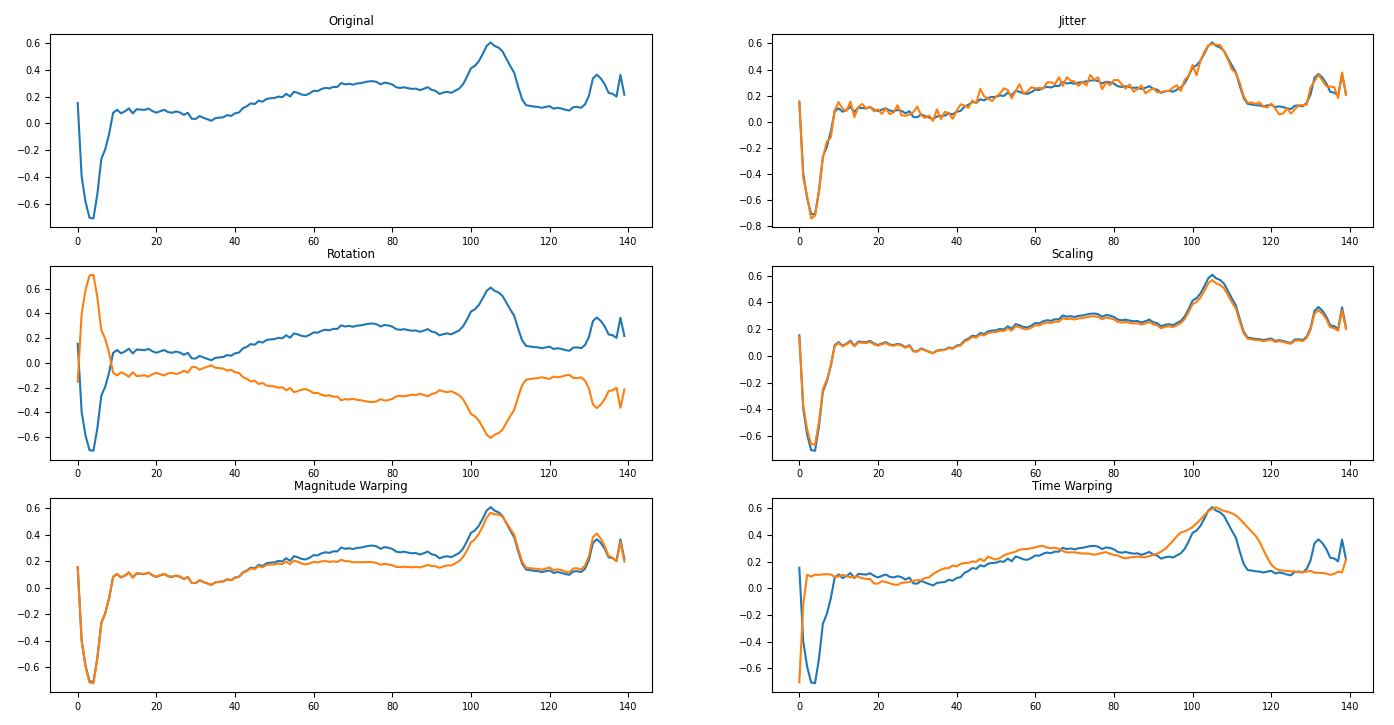}
    \caption{Illustrative Examples of Transformation-Based Methods on the ECG5000 Dataset (Generated instances are depicted in orange)}
    \label{fig:Transformation-Based Methods}
\end{figure}

\subsubsection{Magnitude Transformations}
Magnitude transformations alter the amplitude of time-series data, an approach especially pertinent in contexts where the data's magnitude is informative, such as signal processing or audio analysis. 

For instance, the study by Terry et al. \cite{um2017data} delves into data augmentation techniques like Jittering, Scaling, Rotation, Magnitude-Warping, etc, applied for Parkinson's disease movement classification using wearable sensor data. 
Research underscores that an amalgamation of DA methods markedly propels classification efficacy, jumping from a baseline of 77.52\% to 86.88\%. 
This finding signals a promising trajectory for wearable sensor data augmentation. 
Similarly, Shuang et al. \cite{pi2022improving} harnesses Scaling, Magnitude Warping, and other techniques on I/Q signal data, culminating in a peak recognition accuracy of 93.68\% and successful differentiation between QAM16 and QAM64 signals. 
Additionally, Goubeaud et al. \cite{goubeaud2021white} introduce White Noise Windows (WNW), a novel augmentation technique that fortifies classification methods and mitigates overfitting.

\subsubsection{Time Transformations}
Time transformations adjust the temporal elements of the data, such as data point sequencing or timing. 
They are instrumental when the data's temporal dynamics are crucial, as seen in tasks like time-series classification, forecasting, or sequence prediction. 

In addition to the aforementioned magnitude transformations, the work of Terry et al. \cite{um2017data} also integrates time-based data augmentation techniques such as Permutation and Time-Warping, further demonstrating their utility in enhancing Parkinson's disease movement classification through wearable sensor data. Similarly, the study by Shuang et al. \cite{pi2022improving} employs Time-Warping among its suite of techniques, showcasing its effectiveness in improving the recognition accuracy of I/Q signal data. These instances underline the broad applicability and potential of time transformations in augmenting data for varied classification tasks.

In the realm of construction equipment IMU data, Rashid et al. \cite{rashid2019window} innovates with Window-Warping (WW), a technique that synthesizes training data, thereby enhancing machine learning model proficiency in activity recognition tasks, with K-Nearest Neighbors (KNN) witnessing substantial improvements. 
Furthermore, Guennec et al. \cite{le2016data} discusses two augmentation methods for time series classification with Convolutional Neural Networks: Window Slicing (WS) and Window Warping (WW). 
These strategies, coupled with a semi-supervised Dataset Mixing technique, bolster classification performance, particularly in smaller datasets.

\subsubsection{Frequency Transformations}
Frequency transformations tinker with the time-series data's frequency components, a strategy quintessential in scenarios where the data's frequency content is vital, such as in spectral or vibration analysis. 

In 2023, Yang et al. \cite{yang2023sfcc} pioneers Stratified Fourier Coefficients Combination (SFCC), a method that merges original Fourier coefficients to augment time series datasets, thereby boosting deep neural network performance. 
This efficient technique achieves state-of-the-art results. 
In a similar vein, Goubeaud et al. \cite{goubeaud2021random} unveils Random Noise Boxes (RNB), an innovative augmentation technique optimized for spectrogram classification. 
While particularly potent for ECG and sensor datasets, its efficacy diminishes for motion datasets, underscoring the need for vigilant application to avert performance deterioration.

\subsubsection{Time-Frequency Transformations}
Time-frequency transformations concurrently tweak both temporal and frequency aspects. 
This dual approach is advantageous when both elements are pivotal, enabling the generation of new instances that preserve the original data's temporal and frequency structures. 

In 2019, Lee et al. \cite{lee2019surrogate} employs data augmentation techniques, notably Amplitude Adjusted Fourier Transform (AAFT) and Iterative Amplitude Adjusted Fourier Transform (IAAFT), to concoct surrogate data conducive to deep learning. 
These strategies maintain the time-domain statistics and power spectrum of the original time series, thereby enriching the training of pre-trained Convolutional Neural Networks (CNNs). 
The research corroborates the efficacy of these methods in automated rehabilitation classification tasks, suggesting avenues for further exploration.

\subsection{Pattern-Based Methods}
Pattern-based methods concentrate on discerning and capitalizing on specific patterns or structures inherent in the time-series data. 
These patterns, whether local or global, simple or complex, often involve machine learning models for identification and utilization. 
The hallmark of this category is its emphasis on the data's intrinsic patterns, a focus that proves invaluable in contexts with pronounced structural or pattern-based components, such as electrocardiograms (ECG) or stock market data.

\begin{figure}[!t]
    \centering
    \includegraphics[width=3.5in]{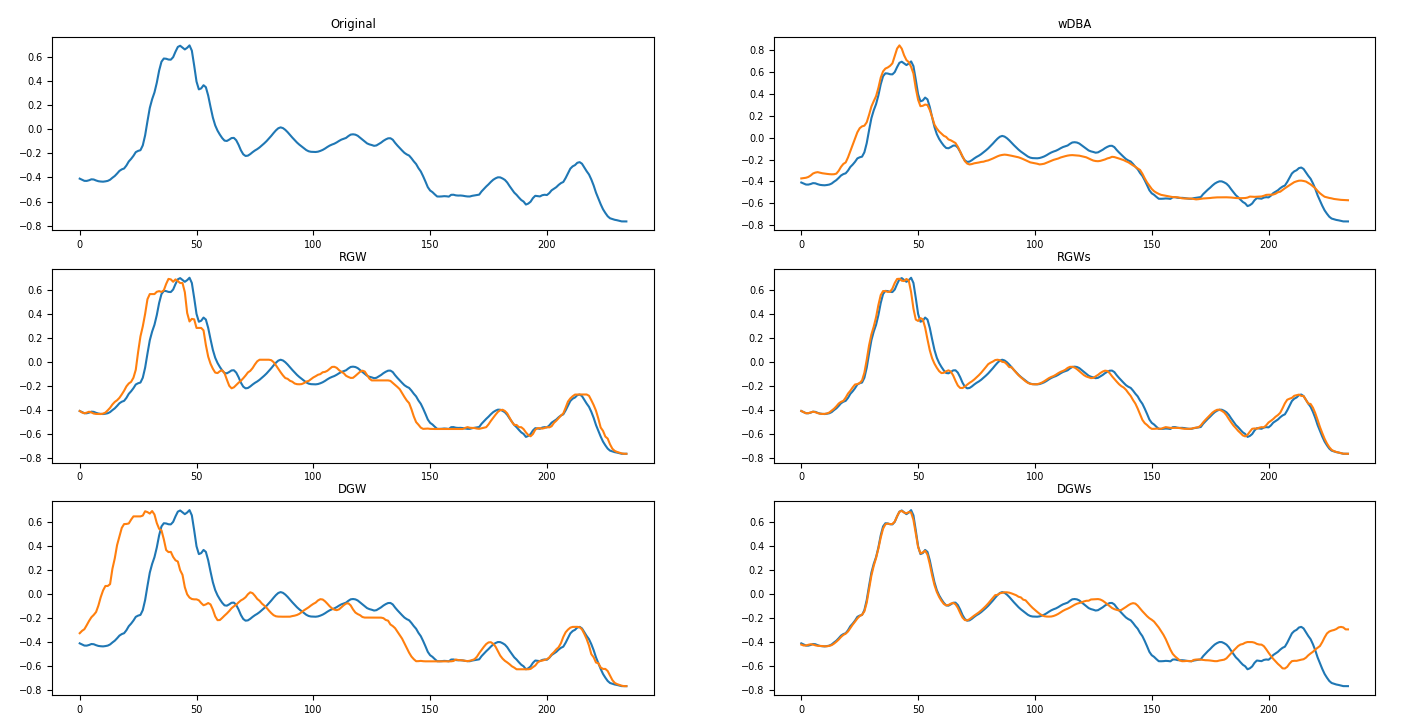}
    \caption{Illustrative Examples of Pattern-Based Methods on the Strawberry Dataset (Generated instances are depicted in orange)}
    \label{fig:Pattern-Based Methods}
\end{figure}

\subsubsection{Magnitude Pattern Mining}
This niche capitalizes on significant local structures or motifs within the data, exploiting these elements to spawn new, realistic data instances. 

In 2019, Kamycki et al. \cite{kamycki2019data} introduces SPAWNER, an innovative data augmentation method that aligns input sequences and restricts the warping path to produce synthetic examples, thereby enhancing the performance of time-series classifiers on benchmark datasets. 
In a parallel study, Aboussalah et al. \cite{aboussalah2022recursive} employs Recurrent In-fill Masking (RIM) in 2022, a novel augmentation method demonstrated to bolster time series classification task performance across synthetic and real-world datasets, outperforming traditional training and other augmentation strategies.

\subsubsection{Time Pattern Mining}
These methods adjust the data's temporal characteristics, maintaining the original data's temporal structure while generating new instances. 

Forestier et al. \cite{forestier2017generating} advocates for a data augmentation method that utilizes Dynamic Time Warping (DTW) to synthesize examples by averaging time series sets, a method especially potent in scenarios with limited example availability. 
The study confirms that augmenting full datasets typically enhances accuracy, with notable improvements in 56 of the 85 tested datasets. 
Similarly, Iwana et al. \cite{iwana2021time} and Cheolhwan et al. \cite{oh2020time} introduce novel methods that leverage time warping and cubic interpolation, respectively, to generate additional training samples effectively, thereby optimizing model performance across various datasets.

\subsubsection{Frequency Pattern Mining}
Frequency Pattern Mining methods focus on manipulating the frequency components of time-series data. 
These methods are pivotal when the data's frequency content is crucial, enabling the generation of new instances that preserve the original data's frequency structure.

In \cite{loris2020paci}, Loris et al. introduce a technique known as EMDA, or Equalized Mixture Data Augmentation, a method premised on spectrograms. 
EMDA operates by forging a new spectrogram through the weighted average of two randomly selected spectrograms from the same class, incorporating perturbations like time delay and equalizer functions. 
This strategy, as per the authors, enhances the diversity and robustness of audio signals, thereby fortifying classification performance. 
Conversely, Cui et al. \cite{cui2015data} advocates for Stochastic Feature Mapping (SFM), a data augmentation method tailored for deep neural network acoustic modeling. 
SFM, a label-preserving transformation, statistically transmutes one speaker's speech data into another's within a specified feature space. 
The essence of SFM is to amplify pattern variations via transformations, thereby bolstering the neural networks' classification invariance and generalization capabilities.

\subsubsection{Time-Frequency Pattern Mining}
Time-Frequency Pattern Mining methods adjust both time and frequency components of the data. 
This dual-pronged approach is beneficial when both the data's temporal dynamics and frequency content are integral, as seen in domains like speech recognition or music analysis.

In 2020, Zhao et al. \cite{zhao2020classification} proposes a unique method that transitions time-domain data into the frequency domain via a discrete cosine transform (DCT), subsequently generating artificial data by merging different frequency bands. 
These synthesized signals are then reverted to the time domain through an inverse DCT, with experimental outcomes underscoring the method's efficacy. 
In a related study, Nanni et al. \cite{nanni2020data} employs data augmentation strategies to enhance the classification accuracy of animal sounds using convolutional neural networks (CNNs). 
These strategies, encompassing transformations of audio signals and their spectrograms, utilize techniques like pitch shift, noise addition, and equalized mixture. 
The findings affirm that data augmentation bolsters the accuracy and generalization capabilities of CNNs, with a fusion of different CNNs yielding optimal results.

\subsection{Generative Methods}
Generative Methods constitute a separate class within the taxonomy of time-series data augmentation techniques. 
These methods leverage generative models to craft new time-series instances, diverging from transformation-based and pattern-based methods by incorporating a learning phase. 
During this phase, the model discerns the data's underlying distribution, subsequently spawning new instances from this distribution. 
This category is invaluable when the objective is to generate diverse, realistic synthetic instances that mirror the original data's statistical attributes.

\begin{figure}[!t]
    \centering
    \includegraphics[width=3.5in]{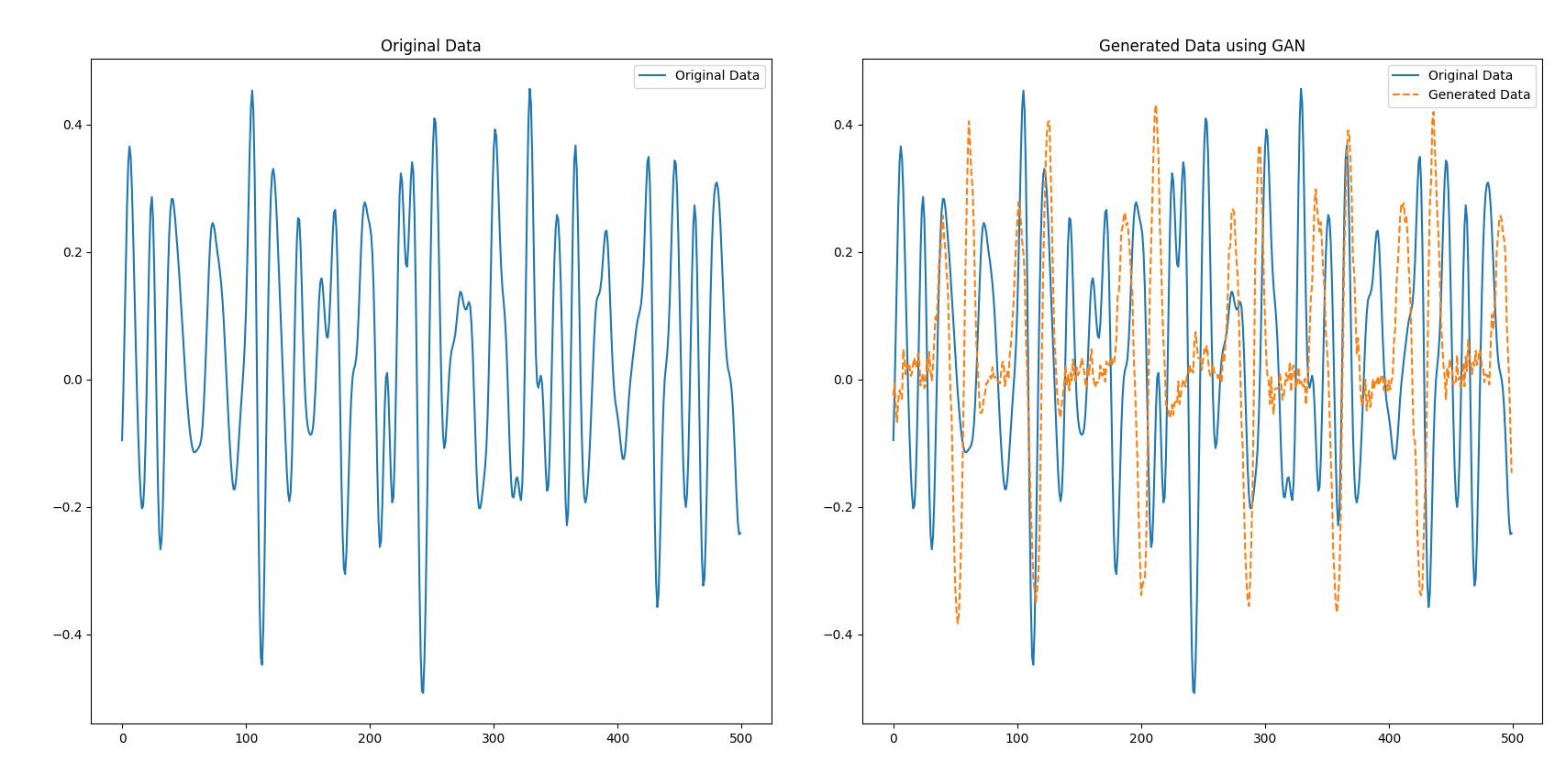}
    \caption{Illustrative Examples of Generative Methods (GAN) on the FordB Dataset (Generated instances are depicted in orange)}
    \label{fig:Generative Methods}
\end{figure}

\subsubsection{Statistical Models}
Statistical Models utilize generative models with pre-established structures to produce new time-series instances. 
These models are advantageous when the data's underlying distribution is known or can be reasonably inferred, facilitating the generation of new instances that adhere to the original data's distribution.

In \cite{cao2014parsimonious}, a Parsimonious Mixture of Gaussian Trees (2MoGT) model is presented by Cao et al., designed for data augmentation in scenarios with imbalanced and multimodal time-series classification. 
This model employs Chow-Liu trees to encapsulate dependencies among time-series features and a Gaussian mixture model to accommodate multimodality. 
It surpasses alternative oversampling methods, especially in datasets with multimodal minority class distributions, and minimizes storage complexity. 
Additionally, Kang et al. \cite{kang2020gratis} introduces GRATIS, a data augmentation technique that generates time series with adjustable characteristics. 
Utilizing a mixture autoregressive (MAR) model, attuned to specific features, it spawns diverse time series. 
The resulting series exhibit diversity, attributable to the mixtures' adaptability and random parameter configurations, proving especially beneficial for analyzing collections of time series with varying lengths, scales, and characteristics.

\subsubsection{Deep Learning Models}
Deep Learning Models employ generative models without predefined forms to create new time-series instances. 
These models are beneficial when dealing with data containing complex, non-linear, or multi-dimensional patterns, as they can generate new instances that encapsulate these intricate patterns.

In \cite{haradal2018biosignal}, a data augmentation strategy based on Generative Adversarial Networks (GANs) is presented by Haradal et al. for biosignal classification. 
This method incorporates a Recurrent Neural Network (RNN) with LSTM units for the generation of time-series data. 
Its efficacy is validated by an uptick in classification accuracy, underscoring its viability for synthesizing biosignals and enhancing biosignal classification. 
Conversely, in 2022, Garcia et al. \cite{garcia2022improving} applies GANs for data augmentation, synthesizing light curves from variable stars. 
This methodology, merged with an innovative resampling technique and evaluation metric, markedly elevates the classification accuracy of variable stars, proving particularly potent for unbalanced datasets and in pinpointing GAN-overfitting scenarios.

Several researchers have also harnessed specialized GANs for data augmentation. 
For instance, Ehrhart et al. \cite{ehrhart2022conditional} employed a conditional Generative Adversarial Network (cGAN) with a diversity term for data augmentation, utilizing an LSTM for the generator and an FCN for the discriminator. 
This method synthesized moments of stress from physiological sensor data, bolstering classifier performance and emerging as a promising technique for stress detection in wearable physiological sensor data. 
Moreover, in 2023, Yang et al. \cite{yang2023ts} introduced a time-series Generative Adversarial Network (TS-GAN) for augmenting sensor-based health data, generating synthetic time-series data that preserves the original data's temporal dependencies, thereby enhancing model performance. 
The study asserts that TS-GAN surpasses conventional augmentation methods in maintaining data consistency. 
Additionally, Deng et al. \cite{deng2022ib} implemented a novel data augmentation strategy, IB-GAN, merging imputation and balancing techniques. 
It boosts classification performance by synthesizing synthetic samples for under-represented classes, averting the mode collapse often seen in standard GANs. 
This approach records substantial performance enhancements against cutting-edge baselines. 
In \cite{chen2019emotionalgan}, an emotion state identification framework and ECG signal augmentation algorithm based on GAN are proposed by Chen et al. 
EmotionalGAN is utilized to synthesize synthetic ECG data, which is then amalgamated with the original training set, enhancing emotion state classification. 
The findings indicate a notable improvement in classification performance with the integration of EmotionalGAN-synthesized data.

Models based on the Encoder Decoder structure have also gained traction in Data Augmentation for Time-Series Classification. 
Moreno et al. \cite{moreno2020improving} employs Variational Autoencoders (VAEs) and Generative Adversarial Networks (GANs) for data augmentation, enhancing prediction accuracy on small datasets, with substantial gains realized through computational effort. 
The study also introduces alterations to balance class representation in synthetic samples. 
Zha et al. \cite{zha2022time} unveils a straightforward yet potent self-supervised model for time series generation, termed the Masked Autoencoder with Extrapolator (ExtraMAE). 
The synthetic time series crafted by ExtraMAE exhibit remarkable practicality and fidelity, eclipsing all benchmarks while remaining more compact and expedient in a minimalist design. 
Furthermore, Hsu et al. \cite{hsu2017unsupervised} introduces two Variational Autoencoder (VAE)-based data augmentation strategies for unsupervised domain adaptation in Automatic Speech Recognition (ASR). 
These methods, encompassing nuisance attribute replacement and soft latent nuisance subspace perturbation, alter nuisance attributes of speech, achieving approximately a 35\% absolute Word Error Rate (WER) reduction.

\subsection{Decomposition-Based Methods}
Decomposition-Based Methods disassemble the original time-series into multiple components, augment each component independently, and subsequently reassemble them to produce new instances. 
This unique tactic of deconstructing the time-series data and individually addressing each component proves advantageous when various components of the data necessitate distinct augmentation techniques.

\begin{figure}[ht]
    \centering
    \includegraphics[scale=0.17]{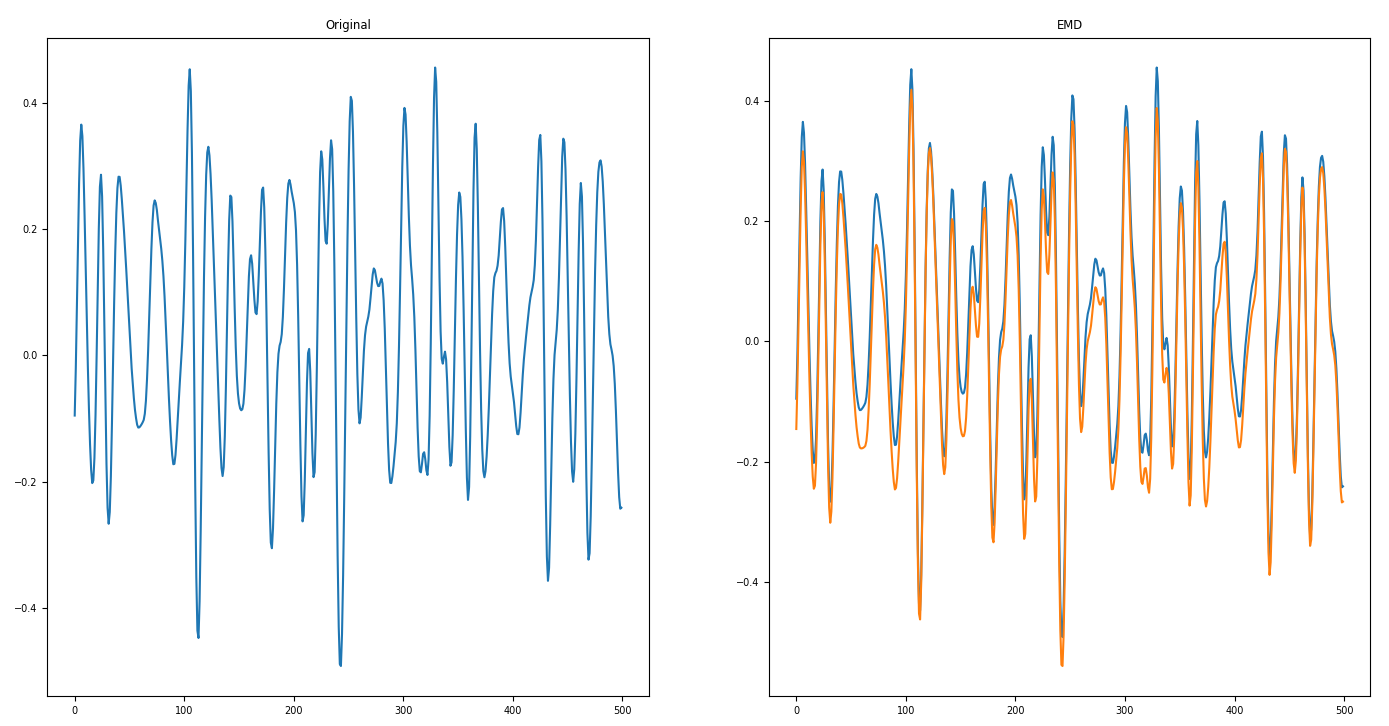}
    \caption{Illustrative Examples of Decomposition-Based Methods (EMD) on the FordB Dataset (Generated instances are depicted in orange)}
    \label{fig:Decomposition-Based Methods (EMD)}
\end{figure}

In \cite{li2022time}, a data augmentation strategy predicated on Empirical Mode Decomposition (EMD) and an Integrated LSTM neural network is showcased by Li et al. 
Raw hand motor signals are fragmented by EMD, followed by the augmentation of the decomposed signals through LSTM. 
The method, trialed on the Ninaweb public dataset, demonstrated a 5.2\% ascension in classification accuracy compared to benchmark outcomes. 
In 2019, Wen et al. \cite{wen2019robuststl} introduces RobustSTL, a pioneering algorithm for time-series decomposition that tackles anomalies, sudden trend shifts, seasonality alterations, and noise. 
It employs bilateral filtering for noise eradication, LAD regression for trend extraction, and non-local seasonal filtering for seasonality determination. 
The algorithm surpasses existing solutions in managing complex time-series data. 
An extension of RobustSTL is presented in \cite{wen2020fast} by Wen et al., designed to accommodate multiple seasonalities in time series data, utilizing a specialized generalized ADMM algorithm. 
This method outperforms leading algorithms like MSTL, STR, TBATS in terms of effectiveness and computational efficiency, with future explorations geared towards multi-resolution seasonal-trend decomposition for further enhancements.

\subsection{Automated Data Augmentation Methods}

Automated Data Augmentation Methods strive to automate data augmentation processes, often discerning the optimal augmentation strategy directly from the data. 
This category distinguishes itself with its focus on automation and adaptability, ideal for circumstances where manual selection and adjustment of augmentation techniques are either impractical or not optimal.

In 2022, Park et al. \cite{park2022dimensional} harnesses PoseVAE and Fast AutoAugment for data augmentation, with PoseVAE rejuvenating skeleton data based on initial human poses, thereby generating stable and novel data. 
Fast AutoAugment fine-tunes the augmentation policy employing RNN-based reinforcement learning. 
The research concludes that these methodologies can effectively augment skeleton data, boosting model performance.

In \cite{fons2021adaptive}, an adaptive weighting scheme for automatic time-series data augmentation is presented by Fons et al., utilizing W-Augment and \textalpha-trimmed Augment methods. 
These techniques amplify the performance of time-series classification tasks, especially in financial markets, by enhancing the accuracy of stock trend predictions. 
The authors ascertain that their methods are computationally efficient and effective on extensive datasets. 
In addition, Cheung et al. \cite{cheung2020modals} introduces MODALS, a modality-agnostic automated data augmentation method operating in the latent space. 

It fine-tunes four universal data transformation operations to conform to various data modalities, creating challenging examples and employing additional loss terms to refine label-preserving transformations. 
It manifests effectiveness across diverse modalities, including text, tabular, time-series, and images.

\section{Experimental Study Setup}
\label{Experimental Study Setup}

In this section, we detail the comprehensive comparative evaluations performed, using various data enhancement techniques. 
These evaluations are designed to provide an empirical comparison between different data augmentation strategies. 
The primary focus of the comparison is to assess the influence of these methods on the performance of neural network models, especially in the context of time series datasets. 
This approach allows for a deeper understanding of how these techniques can be optimized to enhance model performance, offering a broader view of their potential impact across various applications.

\subsection{Datasets}
This research encompasses the application of several time series classification techniques to various datasets from the University of California, Riverside (UCR) time series classification repository \cite{dau2019ucr}. 
The UCR repository is a rich collection of time series datasets, extensively utilized within the academic community for benchmarking time series classification algorithms. 
For this study, fifteen representative categories were chosen, each representing a unique type of time series data.
By including these datasets, we cover all fifteen types of classification datasets within the UCR repository, namely Device, ECG, EOG, EPG, Hemodynamics, HRM, Image, Motion, Power, Sensor, Simulated, Spectro, Spectrum, Traffic, and Trajectory. 
These types encompass a diverse range of domains, such as medical signals, motion capture, sensor readings, and simulated environments, among others. 

The UCR repository is widely recognized for its extensive collection of time-series datasets, each belonging to a distinct category that reflects different real-world applications.
This ensures that our selections are not only classic and representative but also demonstrate the broad applicability and generalizability of our experiments across diverse domains. 
Our goal was to delve deeper into the nuanced relationships between DA methods and datasets. Selecting fewer, but highly representative datasets, allows us to achieve this objective more effectively.

\begin{table}[ht]
\renewcommand{\arraystretch}{1.6}
\caption{Selected datasets from the 2018 UCR Time Series Classification Archive}
\label{tab:datasets}
\centering
\scalebox{0.85}{
\begin{tabular}{l l c c c r}
\hline
\textbf{Dataset} & \textbf{Type} & \textbf{Train} & \textbf{Test} & \textbf{Class} & \textbf{Length} \\
\hline
CBF & Simulated & 30 & 900 & 3 & 128 \\
ECG5000 & ECG & 500 & 4500 & 5 & 140 \\
FordB & Sensor & 3636 & 810 & 2 & 500 \\
GunPointAgeSpan & Motion & 135 & 316 & 2 & 150 \\
ScreenType & Device & 375 & 375 & 3 & 720 \\
Strawberry & Spectro & 613 & 370 & 2 & 235 \\
Yoga & Image & 300 & 3000 & 2 & 426 \\
EOGHorizontalSignal & EOG & 362 & 362 & 12 & 1250 \\
Fungi & HRM & 18 & 186 & 18 & 201 \\
GestureMidAirD1 & Trajectory & 208 & 130 & 26 & Vary \\
InsectEPGRegularTrain & EPG & 62 & 249 & 3 & 601 \\
MelbournePedestrian & Traffic & 1194 & 2439 & 10 & 24 \\
PigCVP & Hemodynamics & 104 & 208 & 52 & 2000 \\
PowerCons & Power & 180 & 180 & 2 & 144 \\
SemgHandMovementCh2 & Spectrum & 450 & 450 & 6 & 1500 \\
\hline
\end{tabular}
}
\end{table}

In Table \ref{tab:datasets}, the term ``Class" refers to the number of distinct categories or labels within the dataset. 
It reflects the granularity and diversity of events or phenomena that the dataset captures. 
A higher number of classes generally suggests a more complex classification task. 
``Length" denotes the duration of each individual time series in the dataset, representing the number of time points or observations for each sequence. 
This factor is crucial as longer sequences might capture more detailed dynamics but also introduce challenges in computational complexity and potential for overfitting.

All datasets were firstly rescaled to ensure that the minimum value in the training set is -1 and the maximum is 1. 
This normalization step helps preserve relative values and maintain consistency across different datasets. 
Additionally, we address missing values, if present, by substituting them with zero, a standard approach that maintains the dimensional integrity of the data while ensuring that missing values do not influence the learning process.

\subsection{Evaluated Network Models}
For our evaluation, we use two deep learning models: the deep convolutional neural network (CNN) model ResNet and the Long Short-Term Memory (LSTM) model. 
ResNet is particularly well-suited for time series data due to its ability to capture hierarchical features through convolutional layers and its robustness to overfitting via residual connections. 
LSTM, on the other hand, excels at learning temporal dependencies and capturing long-term trends in sequential data due to its unique memory cell architecture, which can retain information over long periods.
Both ResNet and LSTM have been widely adopted in the field of data augmentation for time series classification, serving as classic models with extensive usage and validation. 

Many studies have demonstrated their superior performance compared to other models, highlighting their effectiveness in this task.
In our survey, over one-third of the papers utilized ResNet, LSTM, or both to evaluate their proposed algorithms, including studies such as \cite{yang2023sfcc,akyash2021dtwmerge,liu2020efficient,forestier2019deep,oh2020time,do2022data,zhou2020ecg,yang2021time,xu2022mixing,yang2022empirical}. 
This widespread use underscores the robustness and reliability of these models in benchmarking DA methods for TSC.
Our decision to use both ResNet and LSTM stems from their prevalent usage and proven success in many renowned studies. 
Additionally, the unique characteristics of these models make them particularly suited for time series classification. 
By employing these models, we aim to provide a reliable evaluation of DA methods and are confident they will demonstrate excellent results with the methods we have collected.


Hyperparameters for the 1D ResNet used in this evaluation were informed by research conducted by Fawaz et al. \cite{forestier2019deep} and Iwana et al. \cite{iwana2021empirical}, where they noted improvements in time series classification using data augmentation. Similarly, the LSTM configurations were guided by studies \cite{iwana2021empirical,karim2019insights,karim2017lstm} that have shown its effectiveness in handling temporal dependencies in time series data.

\subsubsection{LSTM}
\begin{figure}[ht]
    \centering
    \includegraphics[scale=0.55]{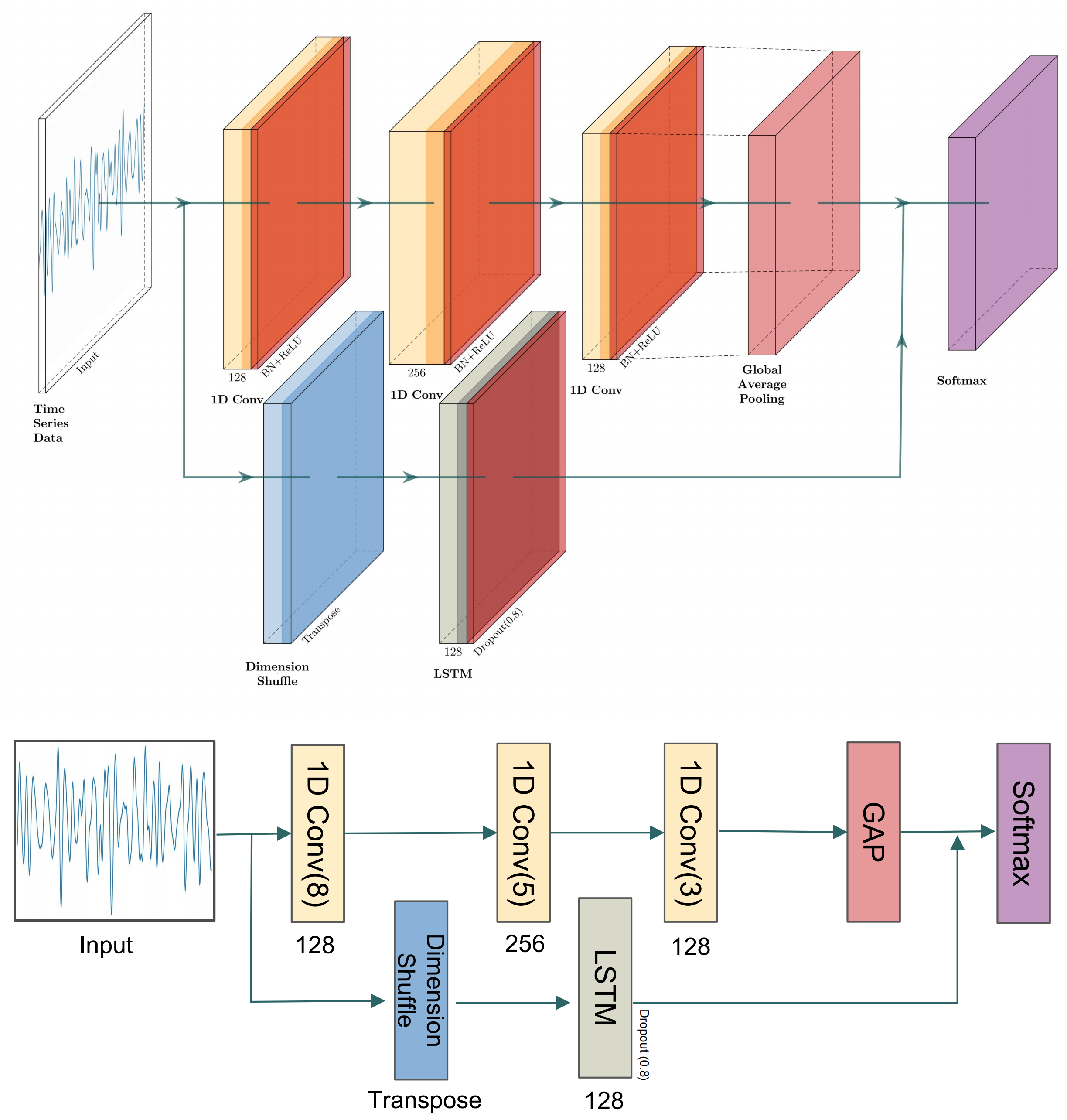}
    \caption{The architectures of the neural networks (LSTM Fully Convolutional Network) used to evaluate the data augmentation methods}
    \label{fig:lstm_fcn}
\end{figure}

As shown in Figure \ref{fig:lstm_fcn}, our implementation of the LSTM network is based on the architecture proposed by Karim et al. \cite{karim2017lstm}. It incorporates both LSTM and convolutional layers to capture temporal dependencies and extract meaningful features from the time-series data. Although this setup uses LSTM in a non-standard way, it has been shown to be more effective than traditional configurations \cite{iwana2021empirical,karim2019multivariate}.

The LSTM component begins by permuting the input shape to align with the multivariate time-step structure required by the LSTM layer. This layer consists of 128 units, followed by a dropout layer with a dropout rate of 0.8, which helps prevent overfitting and enhances generalization.
The convolutional component consists of three 1D convolutional layers. The first layer applies 128 filters with a kernel size of 8 and uses the He uniform initializer for weights. The second layer increases the number of filters to 256, with a kernel size of 5. The third layer reduces the filter count to 128, using a kernel size of 3. Each convolutional layer is followed by batch normalization \cite{ioffe2015batch} and ReLU activation functions, ensuring stable training and promoting non-linearity.
A Global Average Pooling (GAP) layer is used to flatten the output of the convolutional layers, and the results are then concatenated with the output from the LSTM layer. This combination allows the model to leverage both sequential information from the LSTM and spatial features extracted by the convolutional layers.
The final output layer uses softmax activation to produce class probabilities, making the network suitable for multi-class time-series classification tasks.

\subsubsection{ResNet}
\begin{figure}[ht]
    \centering
    \includegraphics[scale=0.49]{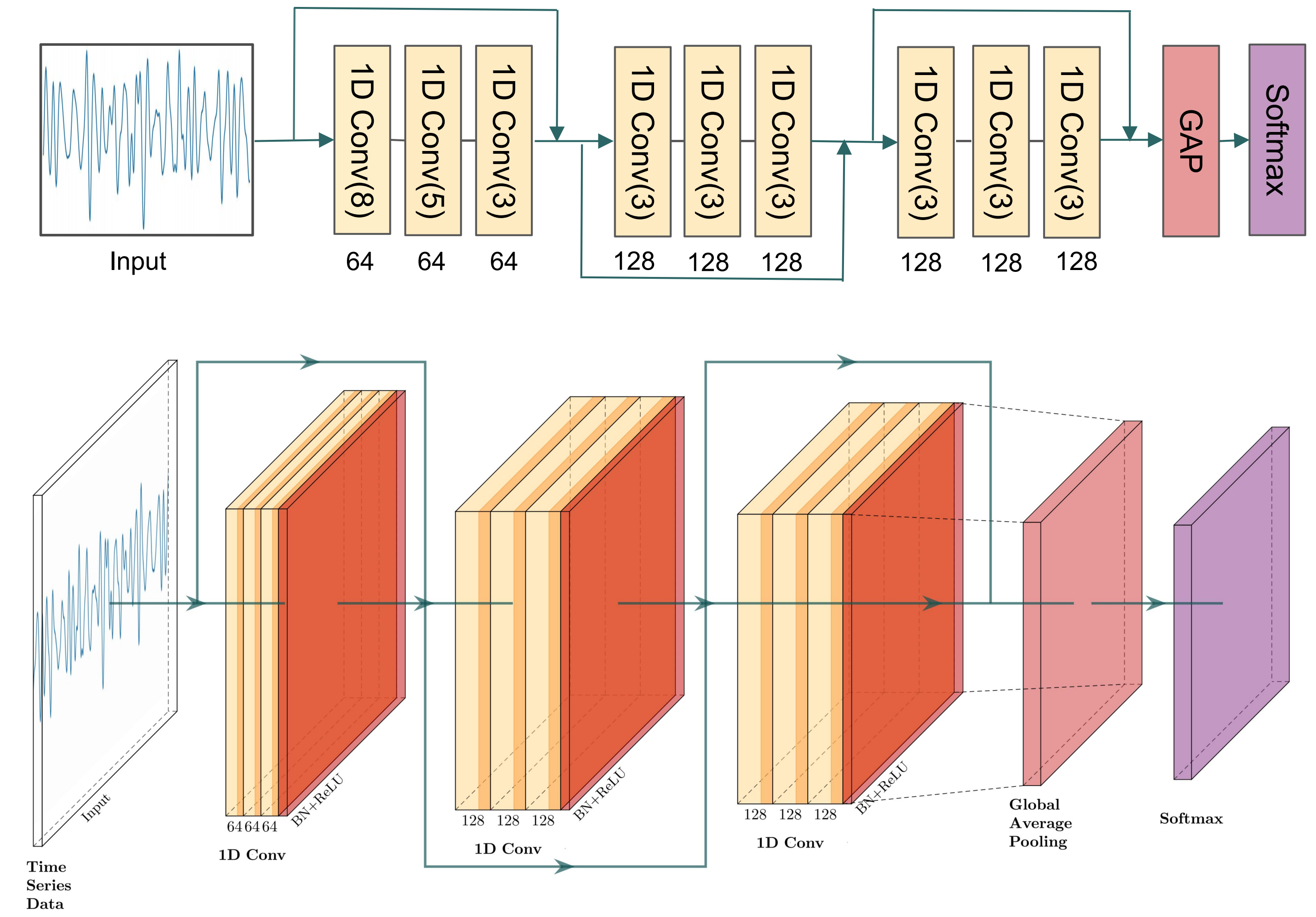}
    \caption{The architectures of the neural networks (ResNet) used to evaluate the data augmentation methods}
    \label{fig:resnet}
\end{figure}

As shown in Figure \ref{fig:resnet}, our implementation of ResNet diverges from the original model \cite{he2016deep} by incorporating only three residual blocks with varying filter lengths and omitting max pooling. 
The first block employs three layers of 64 1D convolutions, the second block contains three layers of 128 1D convolutions, and the third block features three layers of 128 convolutions. 
Each residual block has three convolutional layers with filter sizes of 8, 5, and 3, respectively. 
The batch normalization functions \cite{ioffe2015batch} and the ReLU activation functions follow each convolution. 
Additionally, the residual connections merge the input of each residual block with the input of the subsequent block through an addition operation. 
The final two layers of ResNet comprise a Global Average Pooling (GAP) layer and an output layer with softmax activation.

All network parameters were initialized using the Glorot uniform method \cite{glorot2010understanding}. 
We trained the network using the Adam optimizer \cite{kingma2017adam}, with an initial learning rate of 0.001 and exponential decay rates for the first and second momentum estimates at 0.9 and 0.999, respectively, following Fawaz et al.'s methodology \cite{forestier2019deep}.
To maintain consistency in the comparison of data augmentation methods, we ensured that each model underwent the same 10,000 iterations, reducing the learning rate by 0.1 on plateaus of 500 iterations. 
The number of patterns in the training dataset was quadrupled compared to the original dataset. 
The choice of 10,000 iterations stemmed from observations that indicated that the training loss for the model on each dataset typically converged before this point.

Moreover, as the number of iterations remained constant, the control experiment without augmentation effectively repeated the same patterns to match the augmented experiments. 
Regarding hyperparameter selection, we adhered to the best practices documented in the literature as closely as possible, avoiding excessive adjustments to maintain a fair comparison of the data augmentation effects.

\subsection{Evaluated data augmentation methods}

\begin{table}[ht]
\renewcommand{\arraystretch}{1.3}
\caption{Evaluated Data Augmentation Methods and Their Categories}
\label{tab:da_methods}
\centering
\scalebox{0.90}{
\begin{tabular}{l c c r}
\hline
\textbf{No.} & \textbf{Data Augmentation Method} & \textbf{Category} & \textbf{Reference} \\
\hline
1 & None & - & -\\
2 & Jittering & Transformation-Based & \cite{um2017data}\\
3 & Rotation & Transformation-Based & \cite{um2017data}\\
4 & Scaling & Transformation-Based & \cite{um2017data}\\
5 & Magnitude Warping & Transformation-Based & \cite{um2017data}\\
6 & Permutation & Transformation-Based & \cite{um2017data}\\
7 & Random Permutation & Transformation-Based & \cite{um2017data}\\
8 & Time Warping & Transformation-Based & \cite{um2017data}\\
9 & Window Slicing & Transformation-Based & \cite{le2016data}\\
10 & Window Warping & Transformation-Based & \cite{le2016data}\\
11 & SFCC & Transformation-Based & \cite{yang2023sfcc}\\
12 & SPAWNER & Pattern-Based & \cite{kamycki2019data}\\
13 & wDBA & Pattern-Based & \cite{forestier2017generating}\\
14 & RGW & Pattern-Based & \cite{iwana2021time}\\
15 & RGWs & Pattern-Based & \cite{iwana2021time}\\
16 & DGW & Pattern-Based & \cite{iwana2021time}\\
17 & DGWs & Pattern-Based & \cite{iwana2021time}\\
18 & DTW-Merge & Pattern-Based & \cite{akyash2021dtwmerge}\\
19 & GAN & Generative & \cite{morizet2022pilot}\\
20 & EMD & Decomposition-Based & \cite{nam2020data}\\
\hline
\end{tabular}
}
\end{table}

The data augmentation methods scrutinized in this study are delineated in Table \ref{tab:da_methods}. 
The criterion for their selection was predominantly their recurrent citation in the scholarly literature and the accessibility of their open source implementations, which facilitates reproducibility and comparative scrutiny. 
Amidst a plethora of techniques, some entailing elaborate combinations and complexities, the endeavor was to encapsulate those epitomizing foundational and seminal contributions in the field. 
This strategic choice ensures the study's resonance, relevance, and alignment with contemporary scholarly discourse and benchmarks.
Each method under evaluation was configured using parameters from their respective seminal papers, thereby preserving the authenticity of their original design and ensuring an equitable evaluation landscape. 
These methods were evaluated based on their efficacy in enhancing the performance of both the ResNet and LSTM models across various time series datasets.

We excluded Automated Methods due to their complexity and the need for extensive parameter tuning. 
These methods often integrate multiple algorithms, adding complexity beyond our study's scope, which focuses on foundational algorithms.
The lack of open-source resources also makes reproducibility difficult, potentially compromising reliability. 
Our aim is to evaluate foundational data augmentation techniques that serve as the building blocks for more advanced models, providing a comprehensive understanding of their capabilities and limitations.

Many generated data augmentation techniques require external training and pre-training, adding significant complexity to the evaluation process. 
Each dataset often requires tailored training, making it resource-intensive and challenging to maintain consistency across diverse datasets.
Evaluating these methods also involves careful hyperparameter tuning and managing potential variability in outcomes, affecting reproducibility and reliability.
Due to these challenges and the impracticality of including all methods, we selected GANs as a representative of generated methods, as they offer a manageable balance of complexity within our evaluation framework.

\subsection{Evaluation Metrics}

In addressing the intricate task of evaluating algorithm reliability within the domain of time series analysis, we acknowledge the absence of a singular metric universally applicable across all conceivable applications. 
This context underpins our selection of evaluation metrics, aiming to bridge the gap between general applicability and the specific demands of our study.
This research used a multifaceted evaluation metric system to assess the effectiveness of data augmentation methods. 
Each metric contributes a unique evaluative perspective, cumulatively offering a holistic insight into each method's efficacy. 
The metrics incorporated are as follows:\\

\textit{1. Accuracy (Acc):}
Accuracy, an intuitive and globally recognized metric, gauges the ratio of correctly predicted instances in all prognostications. 
We have opted for Acc as a primary metric due to its relevance in scenarios where the correct identification of both true positives and true negatives is of paramount importance. 
This choice is informed by the nature of our balanced dataset, where equal importance is placed on each class and the distinct types of classification errors are deemed equally significant.
In scenarios where false negatives and false positives are critically important, alternative metrics such as the F1-score might be preferred; however, given the balanced nature of our datasets and the equivalence of misclassification impacts, Acc was deemed the most appropriate metric for our analysis of data augmentation in time-series classification tasks.\\

\textit{2. Method Ranking:}
To facilitate a more nuanced comparative analysis of the data augmentation methods under investigation, we have adopted a ranking paradigm based on accuracy scores. 
This approach not only simplifies the identification of the most effective strategies but also establishes a clear hierarchy of performance across different datasets.
A lower ranking is indicative of superior methodological performance, offering a structured framework to assess the relative success of each data augmentation strategy.\\

\textit{3. Residual Analysis:}
Residual plots were instrumental in juxtaposing the performance of each method with the baseline (None). 
Herein, a residual denotes the deviation between the actual and forecast accuracy. 
Positive residuals signify outperformance over the baseline, whereas negative residuals indicate underperformance. 
These plots are particularly suited to our study as they provide a visual conduit for assessing the relative performance increments or decrements attributable to each data augmentation strategy. 
By visualizing these differences, we can identify patterns and trends in the data that may not be apparent through traditional accuracy metrics alone.

This multimetric evaluation approach underpins the study's commitment to a nuanced, rigorous, and exhaustive assessment, fortifying the conclusions regarding the effectiveness of each data augmentation strategy.

\subsection{Research Questions and Objectives}

Navigating the complexities of time-series data augmentation necessitates a profound understanding of the subtleties and ramifications of various methods. 
This study is predicated on the pursuit of clarity in this domain, aiming to elucidate the efficacy and implications of various data augmentation strategies through the following research questions.\\

\noindent \textit{RQ 1: What is the impact of data augmentation on model performance?}\\

This pivotal question seeks to explore the influence of data augmentation on model efficacy, probing the disparities in outcomes engendered by different techniques. 
It bifurcates into more specific inquiries:\\

\textit{(1) RQ 1.1: What is the extent of the universality of data augmentation strategies?}\\

This inquiry is dedicated to discerning the consistency of data augmentation's effectiveness across various contexts, highlighting any discrepancies and underlying rationales for performance enhancement or degradation.\\

\textit{(2) RQ 1.2: Which data augmentation strategy is the most effective for TSC?}\\

This probe focuses on identifying the preeminent strategy among the evaluated techniques, offering critical insights to practitioners in the time-series domain.\\

\noindent \textit{RQ 2: How do intrinsic dataset characteristics influence the effectiveness of data augmentation and subsequent recommendations?}\\

This exploration delves into the nuanced interplay between dataset attributes and augmentation efficacy, aiming to extrapolate actionable insights for optimal method application based on specific dataset traits. 
The objective is to equip practitioners with informed and customized augmentation strategies, thereby maximizing model performance and circumventing the pitfalls of indiscriminate augmentation application.\\

The ensuing chapter, ``Results Analysis", will systematically address these queries, contributing to a robust, comprehensive understanding of time-series data augmentation and its diverse implications in machine learning.

\section{Results Analysis}
\label{Results Analysis}

In this section, we undertake a methodical examination of the research questions posited at the outset. 
Our first point of inquiry assesses the universality and optimality of various data augmentation techniques (RQ 1). 
Subsequently, we explore the nuanced interdependencies between the attributes of the dataset and the efficacy of these augmentation methods, providing customized guidance for practitioners (RQ 2). 
This discourse contributes a lucid and empirically grounded perspective on the dynamic between data augmentation strategies and time series classification.

\subsection{Answer to RQ 1.1: How universal are data augmentation techniques?}

\begin{table}[ht]
\renewcommand{\arraystretch}{1.3}
\caption{Accuracy Performance of Different Data Augmentation Methods (\%)}
\label{tab:augmentation_performance}
\centering
\scalebox{0.8}{
\begin{tabular}{l c c c r}
\hline
\textbf{Data Augmentation Method} & \textbf{ResNet (\%)} & \textbf{Ranking} & \textbf{LSTM (\%)} & \textbf{Ranking} \\
\hline
RGWs & 85.69 ± 15.12 & 1 & 83.42 ± 17.53 & 2\\
DGWs & 85.65 ± 15.19 & 2 & 82.24 ± 18.63 & 11\\
Window Warping & 85.55 ± 15.32 & 3 & 82.71 ± 18.13 & 6\\
Random Permutation & 85.44 ± 16.02 & 4 & 83.61 ± 16.83 & 1\\
SFCC & 85.43 ± 17.08 & 5 & 83.21 ± 17.38 & 4\\
DGW & 85.32 ± 15.28 & 6 & 82.38 ± 18.94 & 10\\
Permutation & 85.26 ± 16.58 & 7 & 83.24 ± 17.19 & 3\\
RGW & 85.21 ± 15.85 & 8 & 82.61 ± 18.15 & 7\\
DTW-Merge & 85.19 ± 15.67 & 9 & 82.59 ± 18.93 & 8\\
Window Slicing & 85.17 ± 15.62 & 10 & 82.13 ± 18.61 & 12\\
GAN & 85.15 ± 15.22 & 11 & 83.04 ± 18.57 & 5\\
\underline{None} & \underline{84.98 ± 16.41} & \underline{12} & \underline{82.41 ± 18.71}  & \underline{9}\\
wDBA & 84.78 ± 16.36 & 13 & 81.48 ± 20.25  & 13\\
Time Warping & 84.49 ± 15.39 & 14 & 81.47 ± 17.94 & 14\\
Scaling & 84.15 ± 17.27 & 15 & 79.81 ± 23.51 & 15\\
Jitter & 83.95 ± 17.52 & 16 & 78.23 ± 24.55 & 17\\
SPAWNER & 83.26 ± 17.07 & 17 & 76.77 ± 24.95 & 18\\
Magnitude Warping & 83.17 ± 17.75 & 18 & 79.34 ± 23.37 & 16\\
Rotation & 80.21 ± 19.93 & 19 & 76.61 ± 23.46 & 19\\
EMD & 76.57 ± 26.16 & 20 & 73.84 ± 28.69  & 20\\
\hline
\end{tabular}
}
\end{table}

\begin{figure*}[ht]
    \centering
    \includegraphics[scale=0.41]{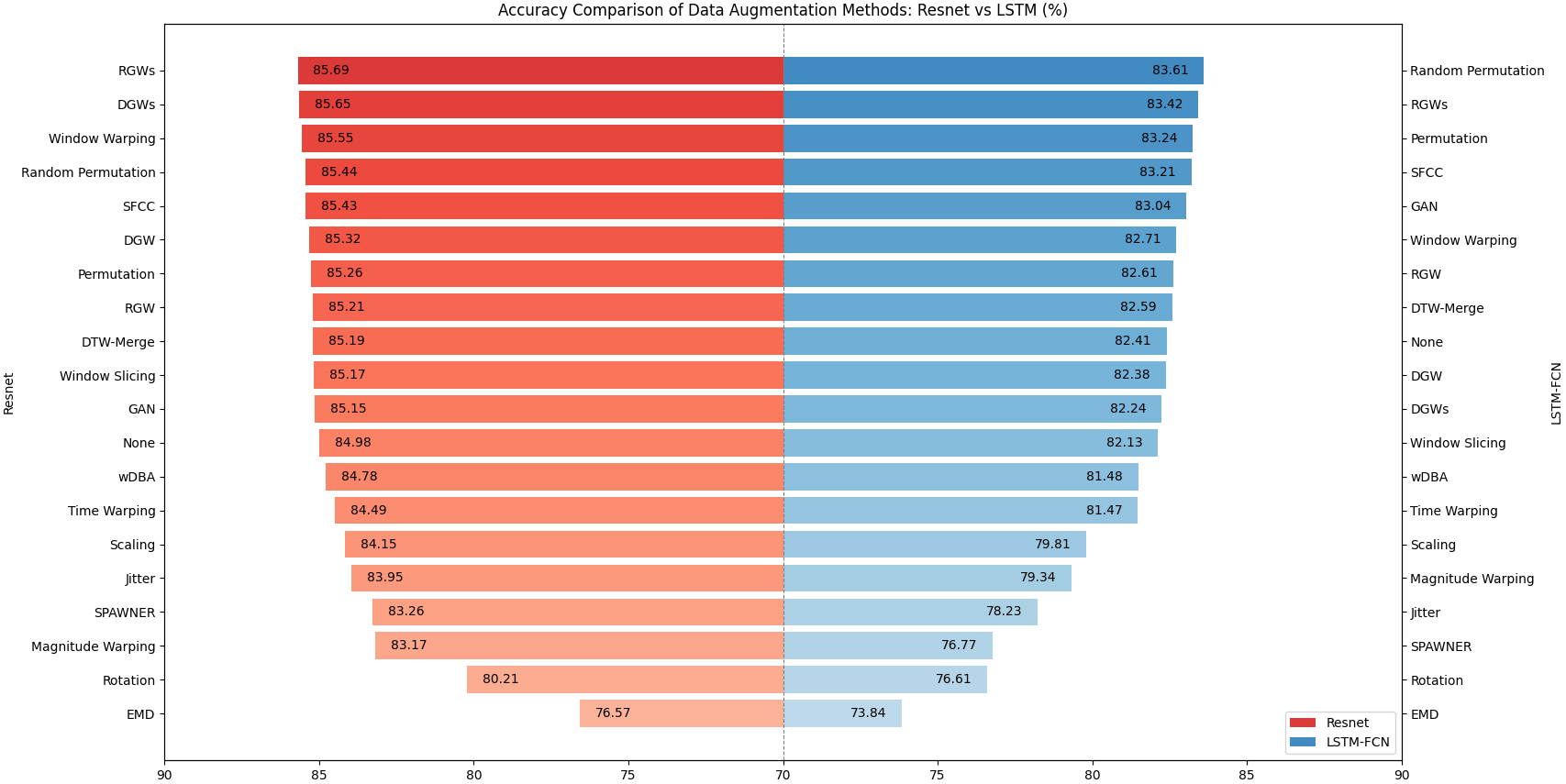}
    \caption{Accuracy Comparison of Data Augmentation Methods: ResNet vs LSTM}
    \label{fig:Performance_of_Different_Data_Augmentation_Techniques}
\end{figure*}

The empirical results of diverse data augmentation strategies are encapsulated in Table \ref{tab:augmentation_performance}. 
In the absence of any data augmentation methodology (labeled ``None''), the model records an average accuracy of $84.98 \pm 16.41\%$ and $82.41 \pm 18.71\%$ on ResNet and LSTM respectively, serving as our performance standard. 
Intriguingly, specific strategies such as Window Warping, DGWs, and RGWs exhibit a slight enhancement in performance compared to this reference point in the ResNet model. 
In the LSTM model, methods such as Random Permutation, Permutation, and SFCC also demonstrate commendable performance. 
Notably, the RGWs method performs exceptionally well across both of the models.
Conversely, certain approaches, notably EMD, incur a significant decrease in efficacy. 

This observation emphatically accentuates that the benefits of augmentation are not universally applicable, emphasizing the necessity for judicious selection of data augmentation techniques. 
While a subset of data augmentation strategies marginally outperforms the established baseline, the distinction in efficacy between several methods and the standard remains negligible. 
For instance, methodologies like Window Slicing, DTW-Merge, and GAN in ResNet show marginal deviations, suggesting that the advantages conferred by some data augmentation techniques may be nuanced and potentially contingent on specific contexts.

\subsection{Answer to RQ 1.2: Which data augmentation technique is optimal for TSC?}

\begin{table}[ht]
\renewcommand{\arraystretch}{1.3}
\caption{Average Ranking of Data Augmentation Methods in each dataset}
\label{tab:av_da_methods}
\centering
\scalebox{0.73}{
\begin{tabular}{l c c c c c r}
\hline
\textbf{No.} & \textbf{Data Augmentation Method} & \textbf{Category} & \textbf{ResNet} & \textbf{Best} & \textbf{LSTM} & \textbf{Best} \\
\hline
1  & Permutation        & Transformation-Based & 7.07 & 2 & 7.33 & 2 \\
2  & Random Permutation & Transformation-Based & 7.93 & 0 & 7.13 & 1 \\
3  & DGWs               & Pattern-Based        & 8.40 & 0 & 9.40 & 1 \\
4  & SFCC               & Transformation-Based & 8.60 & 4 & 7.93 & 1 \\
5  & RGWs               & Pattern-Based        & 9.07 & 0 & 7.13 & 0 \\
6  & \underline{None}   & -                    & \underline{9.11} & \underline{2} & \underline{8.93} & \underline{2} \\
7  & Window Warping     & Transformation-Based & 9.47 & 1 & 8.87 & 0 \\
8  & DTW-Merge          & Pattern-Based        & 9.47 & 0 & 8.93 & 3 \\
9  & DGW                & Pattern-Based        & 10.07 & 0 & 9.80 & 0 \\
10 & Scaling            & Transformation-Based & 10.73 & 0 & 10.20 & 0 \\
11 & RGW                & Pattern-Based        & 10.80 & 1 & 10.40 & 0 \\
12 & Jittering          & Transformation-Based & 11.00 & 0 & 13.27 & 1 \\
13 & GAN                & Generative           & 11.07 & 0 & 7.53 & 2 \\
14 & Window Slicing     & Transformation-Based & 11.40 & 1 & 11.27 & 2 \\
15 & Magnitude Warping  & Transformation-Based & 11.80 & 0 & 10.40 & 0 \\
16 & wDBA               & Pattern-Based        & 12.20 & 0 & 11.87 & 0 \\
17 & Rotation           & Transformation-Based & 12.47 & 2 & 15.27 & 0 \\
18 & Time Warping       & Transformation-Based & 13.13 & 0 & 11.87 & 0 \\
19 & SPAWNER            & Pattern-Based        & 13.20 & 1 & 16.67 & 0 \\
20 & EMD                & Decomposition-Based  & 14.67 & 1 & 15.80 & 0 \\
\hline
\end{tabular}
}
\end{table}

\leavevmode\par

\noindent \textit{1. Transformation-Based Methods}\\

Transformation-Based methods concentrate on subtle modifications to the original data's structure, avoiding the creation of entirely new instances. 
Techniques encompassing Permutation, Scaling, Warping, and Rotation are categorized within this domain.

Referring to Table \ref{tab:augmentation_performance}, both Permutation and Random Permutation diversify the dataset by rearranging time series segments, avoiding extreme modifications that could distort the data's essential structure. 
Similarly, Magnitude Warping and Scaling perform elastic transformations, ensuring that the fundamental data features remain unaltered. 
The success of these strategies reaffirms the criticality of preserving the innate temporal dynamics in time-series data while introducing nuanced variations.

Further experimental results in Figure \ref{fig:Performance_of_Different_Data_Augmentation_Techniques} reinforce the efficacy of the Transformation-Based methods.
In the ResNet model, methods such as Window Warping, Random Permutation, SFCC, Permutation, and Window Slicing all surpass the baseline, with Window Warping, Random Permutation, and SFCC achieving accuracies of 85.55\%, 85.44\%, and 85.43\% respectively, placing them among the top five techniques.
These methods, specifically Window Warping, Random Permutation, and Permutation, belong to the ``Time" branch of Transformation-Based methods, illustrating their capacity to adaptively modify time series data while maintaining its intrinsic properties. 

In the LSTM model, the Transformation-Based methods continue to demonstrate robust performance, with Random Permutation leading at an accuracy of 83.61\%, followed by Permutation and SFCC, all exceeding the performance of the baseline. 
The success of SFCC, categorized under the ``Frequency" branch, highlights the potential of frequency-based transformations to enhance model training without compromising the integrity of the time series data.

These findings indicate that while traditional time manipulation techniques remain effective, incorporating frequency-based augmentations like SFCC can provide a complementary approach to enriching the data representation for time series classification.\\

\noindent \textit{2. Pattern-Based Methods}\\

Pattern-Based methods focus on extracting or amalgamating patterns from the original data, generating new instances that may introduce novel structural elements.

Despite their potential, Pattern-Based Methods have shown mixed performance in our evaluation. 
Remarkably, in the ResNet model, the ``Time" branch of Pattern-Based methods such as RGWs and DGWs achieved the highest accuracies, ranking first and second with scores of 85.69\% and 85.65\% respectively. 
This superior performance underscores the effectiveness of these methods in leveraging temporal patterns without distorting the fundamental time series structure.
Additionally, DTW-Merge also performed above the baseline, further highlighting the potential of Pattern-Based methods to enhance classification accuracy through sophisticated pattern synthesis.

However, the SPAWNER technique from the ``Magnitude" branch, which merges patterns and infuses noise, although creative, has shown limitations, ranking fourth from last in the ResNet model. 
This outcome could be attributed to its approach potentially introducing excessive convolution to the data, which might obscure rather than clarify the underlying patterns in the time series. 

In the LSTM model, the situation is somewhat different.
RGWs again performed exceptionally well, securing the second position with an accuracy of 83.42\%. 
In contrast, methods like DGW and DGWs did not surpass the baseline in the LSTM model. 
This suggests that Pattern-Based methods, which modify sequence structures, may disrupt the continuity that LSTM networks rely on to capture long-term dependencies.
LSTM are sensitive to the order and coherence of time-series data, and altering these patterns could obscure crucial information, leading to reduced accuracy.
This indicates that while effective in ResNet, Pattern-Based methods may not align as well with the sequential learning nature of LSTM architectures.
This variability suggests that these methods might be sensitive to the specific dynamics and characteristics of the dataset, such as inherent correlations or unique motifs, which could be disrupted by the more aggressive transformations employed by some Pattern-Based methods.

The relatively lower performance of SPAWNER, ranking third from last in the LSTM model, corroborates the notion that the synthesis of patterns, while innovative, risks introducing artifacts that deviate too far from the original data configurations, potentially complicating the model's ability to learn the essential features of the data.

These findings suggest a nuanced view of Pattern-Based methods: while they can offer substantial benefits in certain contexts, particularly in models like ResNet that may be better able to handle complex transformations, their application needs to be carefully tailored to the specific characteristics of the data and the model architecture to avoid detrimental effects on performance.\\

\noindent \textit{3. Generative Methods}\\

Generative methods, including models such as GANs, are used to generate new samples that, while structurally divergent from the original data, are designed to be statistically consistent.

In our analysis, GAN, representing Generative Methods, achieved a middle-ground ranking. 
Despite the ability of GANs to generate diverse datasets, the intricacies of time-series data present a formidable challenge, particularly in maintaining temporal coherence while fostering variations.
In the ResNet model, GAN performed only slightly better than the ``None" baseline, ranking just one position higher. 
This modest improvement suggests that while GANs can introduce beneficial variability, the general-purpose GAN model used in our experiment may not be finely tuned to the specific demands of time-series data in the ResNet architecture.

Conversely, in the LSTM model, GAN achieved a more favorable position, ranking 5th with an accuracy of 83.04\%. 
This result could be attributed to LSTM's inherent capability to handle sequential data, where the diversity introduced by GAN might have helped in capturing more nuanced patterns, thus enhancing performance.

However, it is important to note that due to resource constraints, the GAN model we employed was a relatively generic implementation.
The performance of GANs can be significantly influenced by the extent of task-specific training and fine-tuning, particularly for time-series data. 
Specialized GAN models, as highlighted in our taxonomy, have shown promise for time-series generation, but they require extensive optimization to achieve peak performance.
Despite the current limitations, GAN remains a highly innovative and promising method, particularly for its potential in scenarios with limited data availability.\\

\noindent \textit{4. Decomposition-Based}\\

Empirical Mode Decomposition (EMD) is an adaptive method designed to handle nonlinear and non-stationary signals by decomposing them into a finite number of Intrinsic Mode Functions (IMFs) and a residual.
However, in our study, the performance of EMD was suboptimal, ranking the lowest in both the ResNet and LSTM models with accuracies of 76.57\% and 73.84\%, respectively.

Theoretically, EMD's adaptive nature should make it particularly suitable for complex time series signals.
Indeed, in other applications, such as noise classification, EMD has demonstrated effective performance. 
However, its performance in time series classification tasks appears to be limited by several factors. 
Firstly, while the decomposition process can effectively remove high-frequency noise, it may also result in the loss of useful information, especially when critical features are decomposed into higher-order IMFs. 
In our implementation, the strategy of selecting the first two IMFs may lead to information loss in certain cases, thereby affecting the model's performance. 

Moreover, the signals generated by EMD are a combination of multiple IMFs, and the interactions between these IMFs can lead to augmented data that significantly deviates from the original data. 
This deviation could reduce the generalization capability of the classifier. 
Overall, while EMD shows potential in processing complex signals, its performance in time series classification is constrained by several factors.
Future work could explore combining EMD with other signal processing techniques or improving the EMD implementation to enhance its applicability in time series data augmentation.\\

\begin{figure*}[ht]
    \centering
    \includegraphics[scale=0.35]{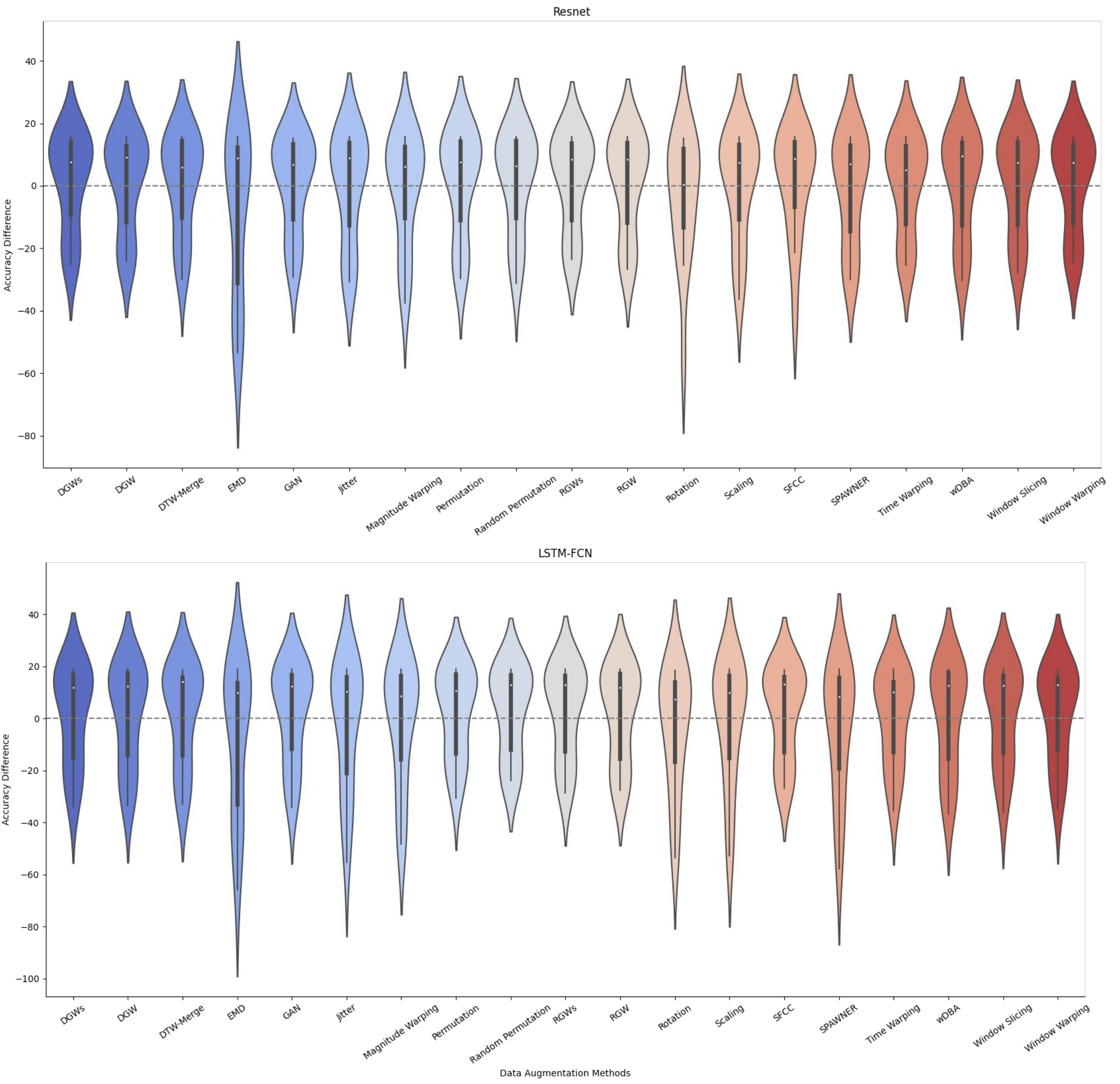}
    \caption{Residual Analysis (Compared to ``None") for each Data Augmentation Method}
    \label{fig:Accuracy_Difference}
\end{figure*}

\noindent \textit{5. Detailed Analysis of Top-Performing and Under-Performing Data Augmentation Techniques}\\

In this subsection, we provide an in-depth examination of the data augmentation techniques that performed exceptionally well, as well as those that underperformed in our experiments. 
This analysis aims to highlight the strengths and limitations of each method, offering insights into their applicability and effectiveness in time series classification tasks.\\

\noindent \textit{(1) Top-Performing Data Augmentation Methods}\\

Our findings, as shown in Table \ref{tab:av_da_methods} of the average ranking and Figure \ref{fig:Accuracy_Difference} of the result's residual analysis, reveal that Permutation, Random Permutation, SFCC, and RGWs surpass our benchmark performance well. 

The consistent presence of these methods at the top is not solely a testament to their effectiveness but also indicative of their inherent capability to maintain and potentially amplify crucial data features during augmentation. 
These methods introduce coherent variations aligned with the underlying data structure, thereby enhancing model training fidelity without straying from the original dataset's essence. 
Their reliability suggests broader applicability across various datasets and experimental conditions and their significance in achieving stable and predictable results.

In particular, SFCC, which ranks fourth overall with an average ranking of 8.60, is notable for its innovation in augmenting time series data by combining stratified Fourier coefficients. 
This method excels in maintaining the integrity of the original data’s frequency components while introducing variability, leading to significant performance improvements.
For instance, SFCC achieved the best accuracy in 4 out of 15 datasets, the highest count among all methods.

RGWs also stands out as a top performer, ranking first and second on average with a strong presence in ResNet and LSTM models. 
This method’s focus on preserving the shape of the time series rather than exact temporal alignment ensures that the augmented data remains true to the original structure, which is critical for time series classification. 

Permutation and Random Permutation, which can be considered as variants of a similar strategy, demonstrate their strength by ranking first and second in average rankings across all the datasets as shown in Table \ref{tab:av_da_methods}, with 7.07 and 7.93 in ResNet, as well as 7.33 and 7.13 in LSTM.
These methods excel at diversifying the dataset by rearranging time series segments in a way that retains the sequence's essential characteristics. 
By doing so, they enhance the model’s generalization capability and reduce the risk of overfitting, which is crucial for achieving robust performance across various tasks.

The success of these top-performing methods underscores the importance of strategies that balance the introduction of variability with the preservation of fundamental data characteristics.
These methods not only improve model accuracy but also offer insights into how different types of augmentations can be optimally applied to various time series classification tasks.\\

\noindent \textit{(2) Under-Performing Data Augmentation Methods}\\

Conversely, Rotation, SPAWNER, and EMD are identified as underperforming data augmentation techniques across the models we evaluated. 
These methods not only exhibited lower accuracy but also demonstrated significant variability in performance, indicating a lack of robustness and generalizability across different datasets.

Rotation and EMD consistently ranked among the lowest-performing techniques in both the ResNet and LSTM models. 
Their substantial fluctuation in performance, as revealed by residual analysis, suggests that these methods might struggle to maintain data integrity across diverse datasets. 
The inconsistent results indicate that these methods may not generalize well and might be more suitable for very specific use cases.
Rotation, for example, could be more effective in the domain of image data, where altering orientation might introduce beneficial variability.
However, in the context of time series classification, the disruption of natural sequence order may obscure temporal dependencies that are critical for accurate modeling. 
Similarly, EMD, which aims to decompose the signal into intrinsic mode functions, might inadvertently disrupt the underlying structure of the time series data, leading to a degradation in model performance.
The high variability observed with EMD further emphasizes its potential instability, possibly due to its sensitivity to the characteristics of individual datasets.

SPAWNER, while slightly outperforming EMD, still exhibits considerable underperformance. 
Its hybrid approach, which merges patterns and introduces noise, may result in overly complex or artificial sequences that do not align well with the natural data patterns. 
This complexity can create challenges for the model in learning meaningful patterns, thus leading to poorer classification outcomes.

These underperforming techniques highlight the importance of selecting appropriate data augmentation strategies that are well-aligned with the nature of the time series data. 
Their limited success underscores the necessity of careful consideration and potentially more sophisticated design or tuning when applying these methods in time series classification tasks.\\

In response to RQ 1 on the universality and optimality of data augmentation techniques for time series classification: The performance of data augmentation methods is diverse, with RGWs and Random Permutation emerging as consistently strong performers. 
These methods have proven effective across various datasets, with SFCC also showing notable performance, particularly excelling in four datasets where it achieved the best results. 

On the other hand, techniques like Rotation, SPAWNER, and EMD significantly underperformed, with high variability and poor generalization across different datasets. 
While some of these methods offer robust improvements in certain dataset, others require more cautious application, often needing to be tailored to specific dataset characteristics to be effective.

\subsection{Response to RQ2: Influence of Dataset Characteristics on Augmentation Effectiveness and User Recommendations}

The realm of time series classification presents perennial challenges, particularly given the heterogeneous nature of datasets. 
Our experimental phase has shed light on the nuanced responses of various UCR datasets to different augmentation techniques. 
This discussion aims to unravel the complex interplay between the attributes of the dataset and the efficacy of the augmentation strategies.\\

\begin{figure*}[ht]
    \centering
    \includegraphics[scale=0.3]{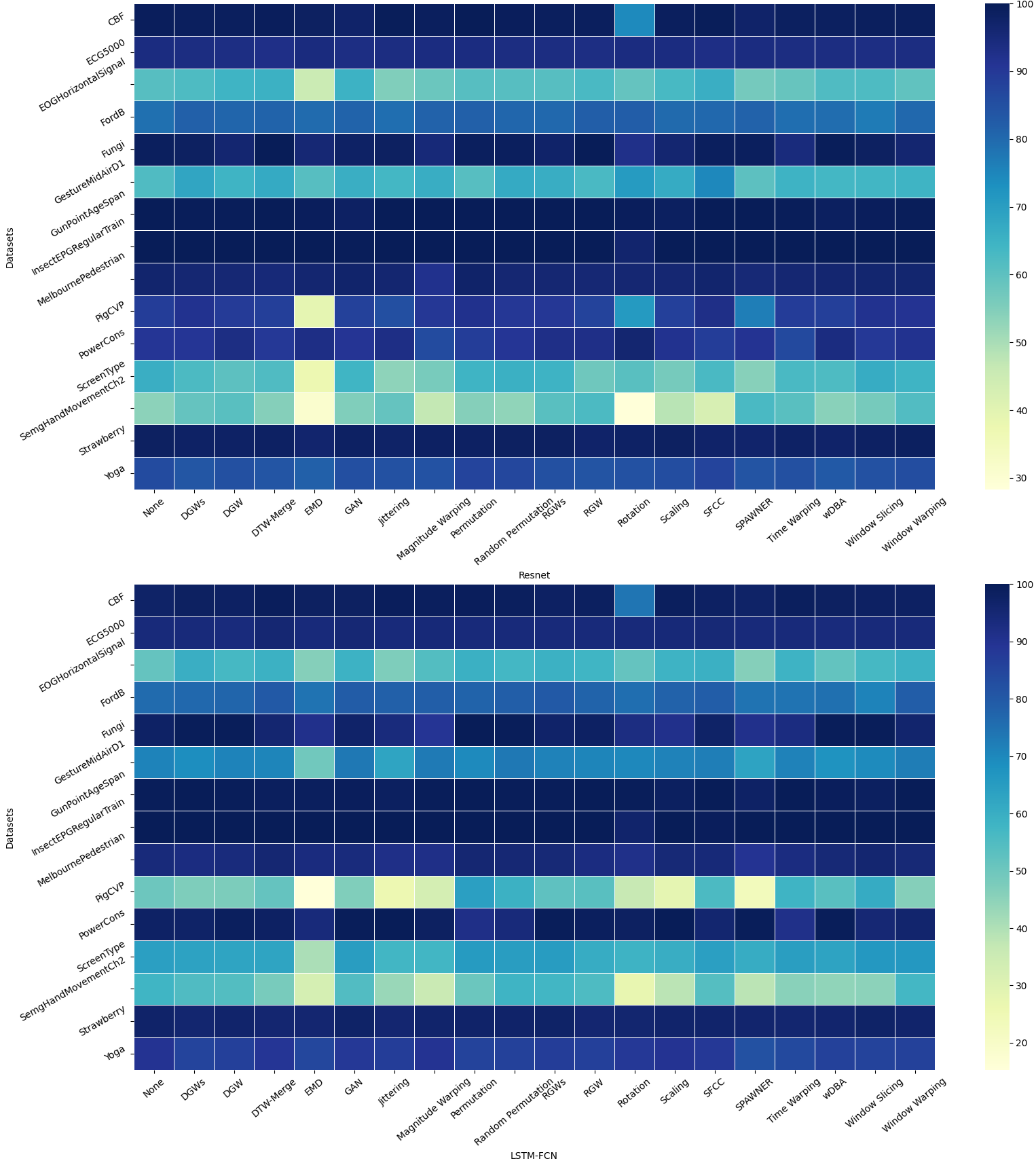}
    \caption{Accuracy Heat Map of Data Augmentation Techniques across all the Datasets}
    \label{Accuracy_of_DA_Techniques_across_Datasets}
\end{figure*}

\noindent \textit{1. Dataset Characteristics Analysis}\\

This subsection delves into the impact of specific dataset characteristics on the performance of different data augmentation techniques, offering insights into how these factors influence the effectiveness of various methods.\\

\noindent \textit{(1) Length of Time Series}\\

Datasets with extended time series, such as PigCVP, with the longest time series among the three datasets, reveals a notable difference in data augmentation method performance between the ResNet and LSTM models. 
In the ResNet model, most data augmentation methods, including SFCC, DTW-Merge, and Window Warping, outperform the baseline, with SFCC achieving the highest accuracy at 92.98\%. 
For longer time series, ResNet benefits from data augmentation techniques that introduce variability while preserving temporal coherence. 
On the other hand, the LSTM model shows a more mixed performance.
While Permutation and Random Permutation significantly outperform the baseline, methods like DTW-Merge and SFCC still surpass the baseline but with a smaller margin. 

A similar trend is observed in the SemgHandMovementCh2 dataset, where ResNet again leverages data augmentation methods effectively, but LSTM struggles more. 
Despite methods like RGWs and Random Permutation outperforming the baseline, others such as SFCC and DTW-Merge fall short, likely due to LSTM's sensitivity to sequence disruptions in longer, complex time series. 
This suggests that LSTM, which heavily relies on sequential data, might be less robust to disruptions in temporal coherence introduced by certain data augmentation methods, especially in long time series.

The MelbournePedestrian dataset, with its short time series, allows various data augmentation methods to perform consistently well. 
The reduced length of the sequences likely diminishes the risk of overfitting, enabling data augmentation methods to introduce variability effectively while maintaining the integrity of the data structure. 
Consequently, both models achieve high accuracy, reflecting the ease with which temporal coherence can be preserved in shorter sequences.

The performance differences across these datasets highlight the significant impact of time series length on the effectiveness of data augmentation methods. 
Longer time series, like those in PigCVP and SemgHandMovementCh2, benefit more consistently from data augmentation methods due to the increased complexity and variability these methods introduce. 
Conversely, shorter time series, such as in the MelbournePedestrian dataset, show more uniform benefits across data augmentation techniques, suggesting that time series length is a crucial factor in determining the effectiveness of data augmentation strategies.\\

\noindent \textit{(2) Dataset Type and Inherent Noise}\\

The performance of different datasets in ResNet and LSTM models reveals important insights into how the type of dataset and inherent noise characteristics influence the effectiveness of data augmentation methods.

As shown in Table \ref{tab:dataset_performance}, datasets with highly regular patterns, such as GunPointAgeSpan and CBF, generally achieve high classification accuracy across both models, often exceeding 97\%. 
These datasets benefit from their structured nature, which contains clear, distinguishable patterns with minimal noise.
The predictability and consistency in these datasets make them less vulnerable to disruptions caused by data augmentation methods.

In contrast, datasets derived from biometric or health-related domains, such as ECG5000 and EOGHorizontalSignal, present a more complex challenge. 
While ECG5000 still maintains high accuracy due to its regularities, EOGHorizontalSignal struggles, particularly in the LSTM model, where accuracy significantly drops. 
This discrepancy is likely due to the higher noise levels and variability inherent in these datasets, which can complicate the model's ability to capture consistent temporal patterns, particularly for methods that rely heavily on sequence integrity.

FordB and ScreenType datasets, representative of sensor-based and device-collected data, exhibit moderate performance across the models. 
Sensor data, by nature, can be prone to noise and irregularities, which may obscure the underlying patterns that models need to learn. 
The moderate accuracy observed in these datasets suggests that the inherent variability and potential noise from sensors challenge the models, making it harder to maintain high accuracy even with the application of data augmentation methods.

SemgHandMovementCh2 and GestureMidAirD1, datasets involving motion or trajectory data, show varied performance, particularly in LSTM, where SemgHandMovementCh2 records notably low accuracy. 
Motion and trajectory data typically involve complex temporal and spatial dependencies, which can be difficult for models to process effectively, especially when disrupted by noise or sequence alterations.
This complexity and potential noise make these datasets challenging for accurate classification.

Lastly, the MelbournePedestrian dataset, classified under traffic data, consistently shows high performance in both ResNet and LSTM models. 
The regular temporal patterns present in traffic data, due to the structured nature of pedestrian movement, likely contribute to this high accuracy.
Despite the potential introduction of noise, the strong and consistent underlying signal in this dataset enables models to maintain high accuracy, highlighting the importance of regularity in the data for successful application of data augmentation methods.

The characteristics of the dataset, such as regularity, noise levels, and temporal complexity, play a crucial role in determining the success of data augmentation techniques. 
Datasets with structured, predictable patterns generally allow for better performance across different models and data augmentation methods, while those with high variability and noise require more careful consideration in the application of augmentation strategies.\\

\begin{table}[ht]
\renewcommand{\arraystretch}{1.6}
\caption{Average Accuracy Performance of Selected Datasets from the 2018 UCR Time Series Classification Archive}
\label{tab:dataset_performance}
\centering
\scalebox{0.75}{
\begin{tabular}{l c c c r}
\hline
\textbf{No.} & \textbf{Dataset} & \textbf{Type} & \textbf{ResNet Accuracy} & \textbf{LSTM Accuracy} \\
\hline
1 & InsectEPGRegularTrain & EPG & \textbf{99.82\%} & 99.77\% \\
2 & GunPointAgeSpan & Motion & \textbf{99.61\%} & 99.16\% \\
3 & PowerCons & Power & 91.14\% & \textbf{97.13\%} \\
4 & CBF & Simulated & \textbf{97.67\%} & 97.10\% \\
5 & Strawberry & Spectro & \textbf{97.87\%} & 96.34\% \\
6 & Fungi & HRM & \textbf{97.55\%} & 95.80\% \\
7 & ECG5000 & ECG & 93.68\% & \textbf{93.98\%} \\
8 & MelbournePedestrian & Traffic & \textbf{95.94\%} & 93.48\% \\
9 & Yoga & Image & 84.95\% & \textbf{86.68\%} \\
10 & FordB & Sensor & \textbf{80.39\%} & 77.03\% \\
11 & GestureMidAirD1 & Trajectory & 65.28\% & \textbf{69.16\%} \\
12 & ScreenType & Device & 60.22\% & \textbf{61.61\%} \\
13 & EOGHorizontalSignal & EOG & \textbf{60.50\%} & 55.48\% \\
14 & SemgHandMovementCh2 & Spectrum & \textbf{53.24\%} & 47.69\% \\
15 & PigCVP & Hemodynamics & \textbf{85.60\%} & 45.44\% \\
\hline
\end{tabular}
}
\end{table}

\noindent \textit{(3) Number of Classes}\\

The diversity of classes within a dataset also impacts the augmentation options. 
From our analysis, datasets with fewer classes generally exhibit better performance across both of the models. 
This trend is particularly evident in datasets like InsectEPGRegularTrain, GunPointAgeSpan, and Strawberry, all of which contain fewer than three classes and consistently rank among the top three in these models. 
The simplicity in class structure likely allows these models to focus more effectively on distinguishing the limited number of patterns, leading to higher accuracy.

Conversely, datasets with a larger number of classes tend to perform less effectively. 
GestureMidAirD1 and PigCVP, which have the highest number of classes among the datasets analyzed, show notably lower accuracies. Both models struggle with GestureMidAirD1, achieving less than 70\% accuracy, which indicates a difficulty in distinguishing between the numerous and potentially overlapping classes. 
The complexity introduced by a high number of classes may dilute the effectiveness of augmentation techniques, making it harder for models to learn distinct patterns within each class.
Interestingly, the PigCVP dataset presents an anomaly. 
While the LSTM model performs poorly on this dataset, with an accuracy around 45\%, the ResNet model achieves a significantly higher accuracy of approximately 85\% (although ranked 10th). This stark contrast could be attributed to the inherent differences in how these models process temporal data. 
ResNet, with its ability to capture hierarchical patterns through convolutional layers, might be better suited to handle the diverse and complex patterns within the PigCVP dataset, even when the number of classes is high.
In contrast, the LSTM model, which relies heavily on sequential data processing, might struggle to maintain consistency across such a varied and class-heavy dataset, leading to its lower performance.

These findings underscore the impact of the number of classes on model performance and highlight the need for tailored data augmentation strategies that consider the class distribution within the dataset.\\

\noindent \textit{2. Recommendations for Data Augmentation Practice}
\leavevmode\\[1.5ex]
This subsection sketches a pragmatic escalation strategy for data augmentation, advising practitioners to start with lightweight, structure-preserving techniques, adopt richer synthetic methods only when data complexity or scarcity requires them, and reserve resource-intensive or automated pipelines for specialized expert-level scenarios.
\leavevmode\\[1.5ex]

\noindent \textit{(1) Transformation-Based Methods}\\

Transformation-Based methods are advisable for datasets where preserving the inherent data structure is critical. 
These methods maintain the core attributes of the original data while introducing controlled alterations that do not disrupt the fundamental sequence or pattern integrity.

These methods are not only critical for preserving pattern-centric time series data but are also relatively simple and intuitive to apply.
Their operation principles are well-established, particularly in both image and time series domains, where they have gained widespread application. 
These methods are computationally efficient, offering quick execution times with minimal resource consumption, making them an excellent starting point for practitioners.

For users, there are many readily available tools and libraries beyond those presented in this paper, which include these fundamental algorithms, making them easy to implement. 
Based on our experimental results, methods like Permutation, Random Permutation, and SFCC consistently demonstrated strong performance, which showed average accuracy improvements of 3-5\% over the baseline, and should be considered as primary candidates for initial attempts.
However, users should note that certain methods, like Rotation, which are effective in computer vision, may not translate well to time series data due to the nature of sequential dependencies.
Additionally, frequency domain methods may not be suitable for non-periodic time series, which requires careful consideration during application.\\

\noindent \textit{(2) Pattern-Based Methods}\\

Pattern-Based methods are invaluable for datasets that are feature-rich or when the introduction of new structural elements could meaningfully enhance the dataset. 
These methods excel at synthesizing variations that can introduce new, potentially beneficial patterns into the data, offering the model additional perspectives that it might not encounter through Transformation-Based methods alone.

Practitioners should carefully assess the complexity of their dataset before applying Pattern-Based methods. 
For datasets with already rich feature sets, such as those with intricate temporal or spatial structures, these methods can be particularly powerful. 
Techniques like RGWs, which performed in the top 30\% on 80\% of evaluated datasets, and DGWs, which excel at introducing elastic deformations, were shown to enhance performance by enabling models to become more robust to slight variations and distortions in time series data.
However, practitioners should be cautious of overcomplicating simpler datasets with these methods, as adding unwarranted complexity could lead to overfitting or a dilution of the signal's clarity.

They are especially beneficial when the goal is to augment the data with synthetic variations that maintain the underlying structure while offering new insights. 
Given their computational demands, it may be advisable to use these methods selectively, perhaps after initial trials with simpler methods like Transformation-Based techniques.
Furthermore, it is important to note that the effectiveness of Pattern-Based methods can vary greatly depending on the inherent noise and variability within the dataset. 
In noisy datasets or those with highly variable patterns, these methods may need careful calibration to avoid introducing excessive complexity that could confuse the model rather than enhance its learning.\\

\noindent \textit{(3) Generative Methods}\\

Generative Methods, such as GANs and VAEs, are powerful tools for augmenting datasets lacking in volume or diversity. 
These techniques can generate new samples, potentially expanding the dataset significantly. 
However, it's essential to ensure that the generated samples align with the original data's statistical properties.

While Generative Methods can offer substantial benefits by significantly expanding the dataset, their application is not as straightforward or universally applicable as other augmentation techniques.
These methods often require extensive external training and fine-tuning tailored to the specific characteristics of the target dataset.
For instance, in our experiments, we utilized a standard, generalized GAN, which, while effective to some extent, did not achieve the level of performance that might be possible with a more specialized generative model.

For specialized tasks, where the generative model is meticulously designed and trained, these methods can outperform traditional augmentation techniques, providing innovative solutions that can produce results unattainable by simpler methods. 
However, the complexity of Generative Methods cannot be understated.
The generative model itself may require extensive computational resources, prolonged training periods, and sophisticated model design. 
Moreover, the availability of ready-to-use tools for these methods is limited, often necessitating custom implementations that are not as user-friendly or accessible as Transformation-Based or Pattern-Based methods.

Practitioners should consider starting with simpler augmentation techniques and move to generative methods as their understanding of the dataset deepens. 
For specialized tasks, where tailored generative models are crucial, these methods can provide unparalleled results, though they come with higher complexity and resource demands.\\

\noindent \textit{(4) Decomposition-Based Methods}\\

Decomposition-Based Methods, despite its potential, is not as widely used as other methods due to its complexity and specific requirements for the dataset.
Our experiments with a general EMD configuration demonstrated subpar performance, ranking last with average ranks of 14.67 and 15.80, suggesting that this method may not be well-suited for broad application unless there is a particular need or a deep familiarity with methods like EMD.

Given the relatively poor results observed in our study, we recommend practitioners to approach Decomposition-Based Methods with caution.
It may be more suitable for niche applications where the decomposition of signals is critical, but for most time series tasks, simpler and more established methods are likely to yield better results with less complexity.
Unless there is a clear justification for its use, or a strong understanding of how to optimize those for a specific dataset, it may be advisable to prioritize other data augmentation techniques that offer greater flexibility and proven effectiveness across a wider range of tasks.\\

\noindent \textit{(5) Automated Data Augmentation Methods}\\

Automated Data Augmentation Methods often involve combining or stacking multiple basic augmentation techniques to optimize performance. 
These methods are not covered in our experiments, as we focused on evaluating individual methods. 
Automated approaches typically require setting weights and fine-tuning various sub-methods to achieve the best results, demanding a deep understanding of each data augmentation technique. 
Therefore, they may not be suitable for beginners or those who aim to use data augmentation methods without extensive customization.

While these methods hold significant potential, especially for complex and large-scale tasks, they are best employed by users with substantial expertise in data augmentation techniques. 
For those dealing with particularly intricate datasets, the investment in designing and fine-tuning an automated approach may be justified. 
However, for most standard tasks, simpler, well-established methods are likely more practical and efficient.\\

\noindent \textit{3. Navigating Complex Scenarios in Data Augmentation Selection}\\

In addressing complex scenarios encountered by practitioners when selecting appropriate data augmentation methods for time series classification, it becomes evident that multifaceted dataset characteristics introduce additional layers of uncertainty. 
This complexity can sometimes lead to conflicts or confusion when attempting to apply the guidelines provided earlier in our study. 
In such intricate cases, our initial recommendation leans towards the employment of Transformation-Based methods as an initial attempt. 
These methods have demonstrated a robust and generally effective approach accompanied by minimal resource and time consumption across various datasets in our experiments, making them a reliable starting point.

However, when faced with datasets that present unique challenges, such as a high number of classes coupled with short series length, the choice between preserving pattern integrity and capturing class-specific patterns requires careful consideration. 
For datasets where the “Length of Time Series" is a distinctive and critical characteristic, prioritizing this feature may guide the selection of data augmentation techniques that respect temporal dynamics without compromising data integrity.

Practitioners are encouraged to experiment with more advanced strategies, such as DGW for nuanced class differentiation, or even generative methods like GANs for enriching dataset diversity.
It is crucial, however, to assess the inherent traits of the dataset and determine which aspects are most vital for the task at hand. 
For instance, if class distinction is paramount, methods enhancing inter-class variability should be prioritized, whereas datasets with a pronounced “Length of Time Series" might benefit more from techniques that efficiently utilize the available temporal information without distorting the original patterns.

The selection of data augmentation methods in complex scenarios demands a balanced approach, integrating empirical evidence from experiments with a nuanced understanding of the dataset's characteristics. 
By initially leaning on the more conservative, transformation-based techniques and gradually integrating more specialized methods, practitioners can navigate the uncertainties inherent in complex data augmentation landscapes, optimizing their approach to suit the unique demands of their datasets.\\

In response to RQ 2 on the influence of dataset characteristics on augmentation effectiveness and user recommendations: Our analysis shows that the length of time series, the number of classes, and the inherent noise in datasets significantly impact the performance of data augmentation methods, with ResNet and LSTM models responding differently to these factors. 

We recommend that users tailor their augmentation strategies to the specific characteristics of their datasets, favoring Transformation-Based Methods for maintaining data integrity, Pattern-Based Methods for feature-rich datasets, Generative Methods for expanding limited datasets with caution, and Decomposition-Based Methods only when there is specific expertise or necessity. 
Automated Data Augmentation, while powerful, should be reserved for complex, large-scale tasks where fine-tuning and expert knowledge are available.

\section{Limitations and Threats to Validity}
\label{Limitations and Threats to Validity}

Every empirical investigation, regardless of the meticulousness in its execution, has certain limitations.
Acknowledging these constraints not only bolsters the study's credibility, but also paves the way for subsequent, more refined research.
In this discourse, we deliberate on the inherent limitations of our study and the strategies used to minimize their implications.

\subsection{Internal Validity Threats}
This subsection scrutinizes potential biases and confounding factors intrinsic to our experimental setup, ranging from dataset coverage and technique breadth to hyper-parameter choices and model dependence—and explains the safeguards we adopted to curb each risk. 
Clarifying these internal constraints helps delineate the conditions under which our empirical claims remain sound.\\
\subsubsection{Limitation of Data Sources}

The primary resource for our study was the UCR time series classification archive.
Although this repository is comprehensive and esteemed within scholarly circles, it encompasses merely a fraction of all conceivable time series datasets. 
To counterbalance this limitation, our selection of datasets from the UCR archive was intentionally diverse, spanning various categories of time series data, from sensor readings to medical diagnostic data, contains all 15 different types in the database.\\

\subsubsection{Scope of Data Augmentation Techniques}

Our evaluation encompassed several data augmentation strategies, yet the expanse of existing and emerging methods is vast.
In order to address this, our research prioritized methods that have gained recognition and widespread application.
Furthermore, we adopted an inclusive approach by utilizing open-source implementations where feasible, establishing a unified framework conducive to future expansions and adaptations.\\

\subsubsection{Influence of Hyperparameter Configurations}

The efficacy of data augmentation strategies is often dependent on hyperparameter configurations. 
Our adherence to parameters endorsed in the scholarly literature does not exclude the potential for alternative configurations to engender different outcomes.
To counterbalance this limitation, we adopted hyperparameters validated by previous empirical studies, ensuring that our findings are anchored in established methodologies and facilitating comparability with existing literature.\\

\subsubsection{Model Specificity}

Our analyses relied heavily on the LSTM and the ResNet architectures.
The implications of using alternative models remain an open question.
The selection of ResNet as well as LSTM was informed by its documented efficacy in time series classification tasks.
We recognize the plethora of alternative models and advocate for additional studies to determine the impact of various data augmentation techniques across diverse neural network architectures.

\subsection{External Validity Threats}
This subsection delineates the principal factors that could constrain the transferability of our findings to other datasets, tasks, and temporal contexts. 
By foregrounding these external validity threats, we equip readers with a clear framework for judging when and how the reported insights can be responsibly generalized.\\
\subsubsection{Generalizability Concerns}

The applicability of our conclusions beyond the datasets and methods employed in our study may be subject to scrutiny.
In order to address this, our methodological design, characterized by a diverse selection of datasets and a wide range of data augmentation techniques, was geared toward maximizing the generalizability of our findings.
However, we advise prudence in extending these conclusions to substantially dissimilar scenarios.\\

\subsubsection{Temporal Context of Tools and Methodologies}

The machine learning landscape is in a state of perpetual evolution, with constant refinements in tools, methodologies, and best practices.
This study captures the state of the art at the time of writing.
While our approach was exhaustive given the current landscape, impending advancements may necessitate new interpretations or adaptations of our findings.\\

\subsubsection{Potential Publication Bias}

Our literature review may reflect a propensity towards methodologies that have demonstrated success or novelty, inadvertently marginalizing equally potent but underreported techniques.
To counterbalance this limitation, we endeavored for a holistic literature review, although we concede the potential for publication bias.
Therefore, we endorse the exploration of lesser-known methodologies and gray literature in future scholarly endeavors.\\

In summary, our study, despite its inherent limitations, was conducted with a rigorous adherence to scholarly standards.
By candidly addressing potential validity threats, we aspire to furnish a robust platform for subsequent inquiries in the realm of time series data augmentation.

\section{Conclusions and Prospects for Future Research}
\label{Conclusions and Prospects for Future Research}
Building upon the analyses presented earlier, this chapter first distills the principal conclusions drawn from our systematic evaluation of time-series data augmentation, and then sketches promising avenues for future inquiry. 
Together, these discussions furnish researchers with a concise roadmap that bridges current insights and forthcoming challenges in the field.
\subsection{Conclusions}

This investigation embarked on a meticulous examination of data augmentation strategies, specifically curated for time series classification.
Upon an exhaustive retrospection spanning the past 10 years, we discerned a plethora of techniques, accumulating over 60 distinct methods from more than 100 scholarly articles.
These methods were subsequently assimilated into an innovative taxonomy, crafted with a focus on time-series classification.
This taxonomy segregates the techniques into five quintessential categories: Transformation-Based, Pattern-Based, Generative, Decomposition-Based, and Automated Data Augmentation, thereby providing a structured reference point for scholars and industry practitioners.

Our inquiry transcended theoretical postulation and went on to practical assessments.
We applied a variety of nearly 20 prevalent data augmentation strategies to all 15 types categories of datasets derived from the esteemed UCR time series classification archive.
The decision to employ ResNet and LSTM as the evaluative model was strategic, motivated by their documented success in analogous research contexts, thus ensuring solid experimental groundwork.
Our multifaceted evaluation approach, encompassing Accuracy, Method Ranking, and Residual Analysis, yielded insights into the nuanced performance dynamics of each technique.

The experimental phase spanned approximately two months, culminating in a baseline model accuracy of 84.98 ± 16.41\% in ResNet and 82.41 ± 18.71\% in LSTM.
In particular, the RGWs, which ranked top 2 in both of the models, substantially improved the performance of the model, while the EMD method was detrimental.
Other strategies, including Window Slicing and GAN, manifested negligible deviations, underscoring the situational specificity of these methods.
Transformation-Based Methods, in particular, have consistently exhibited their merit, potentially attributable to their finesse in maintaining and subtly modifying the inherent structures of time-series data.

Our analysis, extended across various UCR datasets, revealed that the intrinsic characteristics of each dataset significantly influence the success of augmentation strategies.
Based on our research and experimental findings, we provide detailed, actionable recommendations and method suggestions for researchers.
These insights are designed to equip practitioners with well-informed and tailored augmentation strategies, ensuring they maximize model performance while avoiding the risks associated with indiscriminate application of augmentation techniques.

To contribute tangibly to the field, we have integrated most of these strategies into a comprehensive Python Library.
This utility, engineered for user-friendliness, simplifies the application of diverse augmentation strategies, thereby catalyzing further experimental endeavors and innovations.

In essence, this research presents an integrative perspective on the contemporary landscape of data augmentation for time series classification, combining theoretical frameworks with empirical evidence.
The revelations and resources introduced herein are positioned to catalyze continued progress in this domain, fortifying machine learning models against the challenges posed by data limitations, and enhancing their generalizability and robustness.

\subsection{Future Research Directions}
\label{sec:Future_Research_Directions}

The domain of data augmentation for time-series classification is expansive and continually evolving.
The milestones achieved in this research constitute a foundational layer with expansive potential for further exploration and refinement.
The following are prospective directions for subsequent inquiries.\\

\subsubsection{Exploration of Diverse Neural Network Architectures} 
While the present study was anchored around a select group of established neural network models, the dynamic nature of deep learning architectures intimates the potential advantages of a more extensive exploration.
In particular, the adaptability of emerging models, especially those designed for sequential data like Transformer models, warrants comprehensive investigation.\\

\subsubsection{Extension to an Expanded Dataset Spectrum} 
The scope of our study was limited to a specific selection of datasets, representing only a segment of the myriad applications of time series classification.
Future research could broaden this scope to include datasets from diverse fields such as finance, environmental science, or social dynamics, thus testing the universality and domain-specific nuances of the augmentation techniques.
Additionally, a promising direction is the creation of synthetic datasets with controlled inner characteristics.
For instance, generating time series data from ARMA processes could offer a structured approach to examining how different data augmentation methods influence time-series classification across varied data types. 
By manipulating these synthetic datasets, researchers can gain deeper insights into the specific strengths and weaknesses of each augmentation method, ultimately enhancing their application to real-world datasets.\\

\subsubsection{Advanced Augmentation Selection and Combination} 
Within the domain of data augmentation, the synergistic integration of disparate techniques emerges as a pivotal factor in augmenting the efficacy of model training. 
However, this integration necessitates a methodical selection process to mitigate potential conflicts and eliminate redundancy. Our empirical analysis elucidates that specific technique combinations may either synergize or antagonize, thereby exerting a differential impact on model efficacy. 
Illustratively, the amalgamation of noise injection with geometric transformations might veil essential data patterns, whereas the concurrent application of jittering and scaling could effectively replicate real-world data variabilities. 
This underscores the imperative for a data-driven approach to data augmentation, predicated on a nuanced comprehension of dataset characteristics.
Furthermore, the conduct of preliminary experiments or small-scale pilot studies is posited as a crucial step towards garnering insights into the compatibility and efficacy of various augmentation technique combinations. 
Such investigative endeavors facilitate the elucidation of optimal strategies that not only maximize model performance but also safeguard the integrity of the fundamental information content inherent within the data.

Innovative methods such as RandECG \cite{nonaka2021randecg}, which judiciously integrate multiple enhancements through transparent and interpretable hyperparameters, herald a promising future for augmentation synthesis.
Adapting such strategies for time-series data, potentially with the incorporation of advanced frameworks like reinforcement learning or meta-learning, could optimize augmentation processes.
Furthermore, ensuring the computational efficiency of these algorithms, especially in scenarios with extensive datasets or limited processing capabilities, remains an essential consideration.


%





\ifCLASSOPTIONcaptionsoff
  \newpage
\fi



\bibliographystyle{IEEEtran}
\bibliography{References}
\end{document}